\documentclass[10pt]{article}

\usepackage[paper=a4paper,top=18mm,bottom=20mm,left=20mm,right=20mm]{geometry}
\usepackage{amsmath,amssymb}
\usepackage{fontspec}
\usepackage{graphicx}
\usepackage{booktabs}
\usepackage[table]{xcolor}
\usepackage{caption}
\usepackage{float}
\usepackage{titlesec}
\usepackage{fancyhdr}
\usepackage{enumitem}
\usepackage{multicol}
\usepackage{balance}
\usepackage{ragged2e}
\usepackage[skip=0.55em plus 0.1em, indent=0pt]{parskip}
\usepackage[colorlinks=true, linkcolor=ink, citecolor=g500, urlcolor=ink]{hyperref}

\definecolor{ink}{HTML}{0A0A0A}
\definecolor{g50}{HTML}{F7F7F7}
\definecolor{g100}{HTML}{EEEEEE}
\definecolor{g200}{HTML}{DDDDDD}
\definecolor{g300}{HTML}{BEBEBE}
\definecolor{g400}{HTML}{909090}
\definecolor{g500}{HTML}{6B6B6B}
\definecolor{g600}{HTML}{4A4A4A}
\definecolor{bestcell}{HTML}{EEEEEE}
\definecolor{runnerupcell}{HTML}{F7F7F7}
\newcommand{\best}[1]{\cellcolor{bestcell}\textbf{#1}}
\newcommand{\runnerup}[1]{\cellcolor{runnerupcell}#1}
\arrayrulecolor{g200}

\newfontfamily\inter[Extension=.ttf,UprightFont=Inter-Regular,BoldFont=Inter-SemiBold]{Inter}
\newfontfamily\intermed[Extension=.ttf,UprightFont=Inter-Medium]{Inter}
\newfontfamily\brock[Extension=.otf,UprightFont=Brockmann-SemiBold,BoldFont=Brockmann-Bold,ItalicFont=Brockmann-SemiBoldItalic,LetterSpace=-2.5]{Brockmann}
\newcommand{\eyebrow}[1]{{\intermed\fontsize{7.2}{9}\selectfont\addfontfeature{LetterSpace=14}\color{g500}\MakeUppercase{#1}}}
\newcommand{\eyebrowmono}[1]{{\ttfamily\fontsize{6.8}{9}\selectfont\addfontfeature{LetterSpace=14}\color{g500}\MakeUppercase{#1}}}

\newcommand{\twodigits}[1]{\ifnum#1<10 0\fi\number#1}

\renewcommand{\thesubsection}{\arabic{section}.\arabic{subsection}}
\titleformat{\section}{\brock\fontsize{15.5}{18}\selectfont\color{ink}}{\twodigits{\value{section}}\;·\;}{0pt}{}
\titlespacing*{\section}{0pt}{1.9em plus 0.3em}{0.55em}
\titleformat{\subsection}{\inter\bfseries\fontsize{10.5}{13}\selectfont\color{ink}}{\thesubsection}{0.5em}{}
\titlespacing*{\subsection}{0pt}{1.3em plus 0.2em}{0.3em}
\titleformat{\paragraph}[runin]{\inter\bfseries\color{ink}}{}{0pt}{}
\titlespacing*{\paragraph}{0pt}{0.7em}{0.45em}

\DeclareCaptionLabelFormat{sl}{{\eyebrowmono{#1 #2}}}
\DeclareCaptionFormat{slfig}{\noindent{\color{g200}\rule{\linewidth}{0.4pt}}\par\vspace{4pt}\noindent#1\hspace{0.7em}#3\par}
\DeclareCaptionFormat{sltab}{\noindent#1\hspace{0.7em}#3\par\vspace{3pt}}
\newcommand{\thd}[1]{{\ttfamily\fontsize{6.8}{9}\selectfont\addfontfeature{LetterSpace=14}\color{g500}\MakeUppercase{#1}}}
\renewcommand{\toprule}{}
\renewcommand{\midrule}{\specialrule{1.4pt}{2pt}{4pt}\arrayrulecolor{g200}}
\renewcommand{\bottomrule}{\specialrule{0.4pt}{2pt}{0pt}}

\AtBeginEnvironment{tabular}{\small\arrayrulecolor{ink}}

\fancypagestyle{first}{\fancyhf{}\fancyfoot[L]{\eyebrowmono{Preprint · 2026 · Speridlabs Research}}\fancyfoot[R]{\eyebrowmono{ENEAS · \thepage}}}
\setlist[itemize]{leftmargin=1.2em,itemsep=0.3em,label={\color{g400}\rule[0.5ex]{0.5em}{0.5pt}}}

\newsavebox{\abstractbox}
\newlength{\abstractH}
\renewenvironment{abstract}{%
  \begin{lrbox}{\abstractbox}\begin{minipage}{\dimexpr\linewidth-14.2mm\relax}%
  \vspace{4mm}\setlength{\parskip}{0.5em}\small\linespread{1.22}\selectfont%
  \noindent\eyebrow{Abstract}\par\vspace{0.6em}}%
  {\par\vspace{4mm}\end{minipage}\end{lrbox}%
  \setlength{\abstractH}{\dimexpr\ht\abstractbox+\dp\abstractbox\relax}%
  \noindent{\setlength{\fboxsep}{0pt}\setlength{\fboxrule}{0.4pt}%
  \fcolorbox{g200}{g50}{\raisebox{-\dp\abstractbox}{\color{ink}\rule{2.2pt}{\abstractH}}\hspace{6mm}\usebox{\abstractbox}\hspace{6mm}}}\par\vspace{6mm}}

\newcommand{\preprinttitle}{%
  \setlength{\parskip}{0pt}%
  \noindent\begin{minipage}[b]{0.5\linewidth}\raisebox{-0.35\height}{\includegraphics[height=6.2mm]{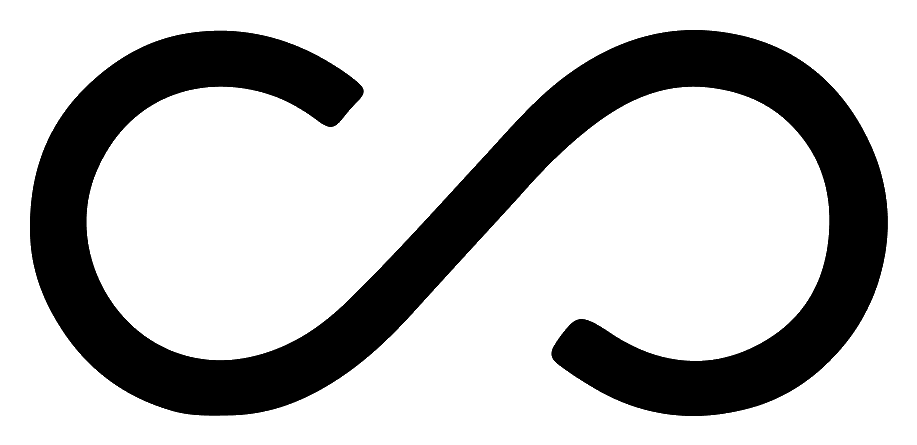}}\hspace{2.2mm}{\inter\fontsize{12}{14}\selectfont\color{ink}speridlabs}\end{minipage}%
  \begin{minipage}[b]{0.5\linewidth}\raggedleft\eyebrowmono{Preprint · Technical report}\\[2pt]\eyebrowmono{2026 · Speridlabs Research}\end{minipage}\par
  \vspace{3mm}\noindent{\color{ink}\rule{\linewidth}{1.4pt}}\par\vspace{7mm}%
  \noindent\eyebrow{Text-promptable segmentation}\par\vspace{2.5mm}%
  \begin{flushleft}\brock\fontsize{25}{27}\selectfont\color{ink}ENEAS: Embedding-guided Neural Ensemble for Adaptive Segmentation.\end{flushleft}\vspace{2.5mm}%
  \begin{flushleft}\inter\fontsize{10.5}{14}\selectfont\color{ink}%
  \textbf{Javier del Pino}\textsuperscript{\textdagger}, \textbf{Salvador Rodríguez}, \textbf{Alejandro Garabito}, \textbf{Javier Álvarez}, \textbf{Chema Garabito}\\[0.5em]
  {\footnotesize\color{g500}SperidLabs\;·\;\textsuperscript{\textdagger}Project lead\;·\;Code and models: \href{https://github.com/speridlabs/eneas}{\color{ink}github.com/speridlabs/eneas}}\end{flushleft}\vspace{2mm}%
  \setlength{\parskip}{0.55em plus 0.1em}}

\makeatletter
\renewcommand\@biblabel[1]{{\ttfamily\fontsize{7}{9}\selectfont\color{g500}[\ifnum#1<10 0\fi#1]}}
\renewenvironment{thebibliography}[1]{%
  \section*{References}\addcontentsline{toc}{section}{References}%
  \fontsize{8}{10.5}\selectfont\color{g600}%
  \list{\@biblabel{\@arabic\c@enumiv}}{\settowidth\labelwidth{\@biblabel{#1}}%
  \leftmargin\labelwidth\advance\leftmargin\labelsep\itemsep 0.25em\parsep 0pt\usecounter{enumiv}}%
  \sloppy\clubpenalty4000\widowpenalty4000}{\endlist}
\makeatother

\begin{document}
\preprinttitle\thispagestyle{first}

\begin{abstract}
We present ENEAS, a unified, text-promptable method for instance tracking and semantic discovery.
Text-promptable segmentation models, including the latest foundation models such as
SAM 3~\cite{sam3}, still suffer from
temporal hallucinations, spatial fragmentation, and semantic misclassification: they fail to report
target absence when an object leaves the field of view, segment local textures instead of the
complete object during extreme close-ups, and prioritize visual features over ontological reality,
so that visually similar artifacts such as statues, paintings, or reflections are segmented as
target entities.

ENEAS works two ways from a single method: precise tracking and high-quality segmentation of a
unique instance, and open-concept discovery of every instance a text query names, resolved by a
semantic verification layer. For tracking, we extend the geometrically robust SeC architecture,
previously limited to point interactions, with a text-prompting adapter and leverage its temporal
memory, so that the target is held through disappearance without drifting to distractors and kept
whole even when it fills the entire view. For discovery, the verification layer combines
high-speed visual embedding matching with conditional VLM refinement, invoking semantic reasoning
only for ambiguous candidates, which filters out the ontological errors that visual-only models
cannot distinguish while keeping latency low. Designed with 3D reconstruction in mind, where a
single misclassified distractor corrupts the asset, ENEAS unlocks high-quality semantic tracking
and segmentation of video, of broad libraries, and of collections of temporally or spatially
unordered data, together with the discrimination to tell true instances from their doppelgangers:
things that look alike but are not the same. The code and models are available at
\href{https://github.com/speridlabs/eneas}{github.com/speridlabs/eneas}.
\end{abstract}

\noindent\includegraphics[width=\linewidth]{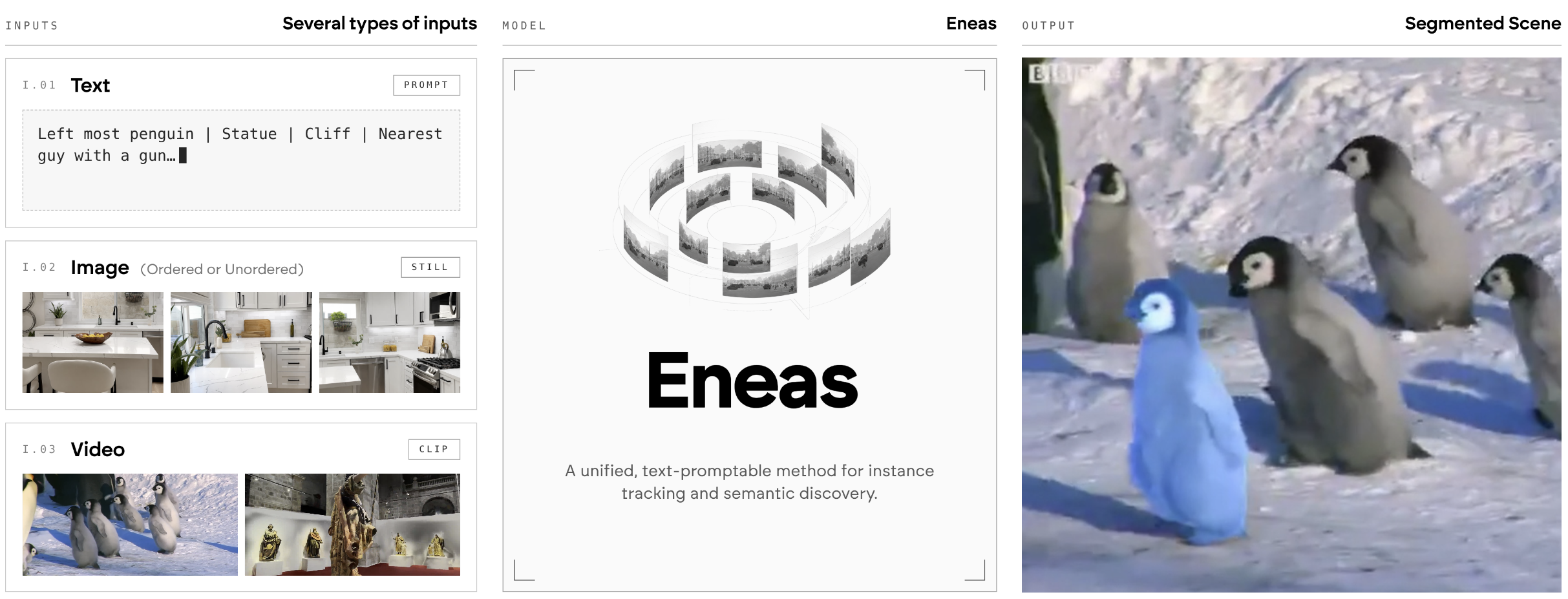}\par\vspace{5mm}
\begin{multicols}{2}

\section{Introduction}

Text-guided segmentation has become a central primitive of modern visual understanding. A user
names what they care about, and a model is expected to return pixel-accurate masks for it, either
following one specific object across a video or discovering every instance of a concept as the
scene evolves. Foundation models such as SAM~\cite{sam}, SAM 2~\cite{sam2} and, most recently,
SAM 3~\cite{sam3}, together with open-vocabulary detectors~\cite{groundingdino,florence2} and
multimodal language models~\cite{llava,qwen3vl}, have made both capabilities broadly accessible, unifying promptable tracking and
open-vocabulary discovery in a single family of models.

Despite this progress, two failure modes persist on uncurated video. Detection-driven trackers
lack a notion of identity: when the target disappears they re-detect the prompt on whatever looks
similar, and under extreme close-ups they fragment it into its parts. Semantic discovery, in
turn, resolves ``what is this'' through visual priors alone, so a hyper-realistic statue or a
pop-art portrait is segmented as a person. The segmentation is geometrically correct and
semantically wrong.

We argue that these failures have a common root: strong perception coupled with weak verification.
We met them while building 3D reconstruction pipelines, where a specific object must be isolated
across unordered, partially occluded views, and where every visitor must be removed without
touching the sculptures being reconstructed. Existing models such as SAM 3 returned several
paintings when asked for one, or masked statues as people, and were not usable for the task.
This paper presents \textbf{ENEAS} (Embedding-guided Neural Ensemble for Adaptive Segmentation), a
unified method that closes this gap. It takes a natural-language prompt and a sequence of frames,
which may be an ordered video or an unordered image collection, and returns binary masks. When the
prompt refers to a specific instance, the method anchors it once on a reference frame and propagates it
through time with a memory-based tracker that maintains identity and reports absence. When the
prompt names a category, it re-evaluates every frame, proposes candidate regions, scores them
with a vision-language embedding, and routes only the ambiguous ones to a vision-language model
(VLM) acting as a semantic judge. Both behaviours share the same grounding and segmentation
components and are exposed through the same interface.

Our contributions are the following:
\begin{itemize}
\item \textbf{A unified text-guided segmentation method} that handles instance tracking
  and semantic discovery within one framework, applicable to ordered video and unordered image
  collections alike.
\item \textbf{Text-driven initialization for memory-based tracking.} We extend the SeC
  tracker~\cite{sec}, previously limited to point interaction, with an open-vocabulary grounding
  stage, obtaining a text-prompted tracker that handles disappearance without drifting and preserves
  spatial integrity under extreme scale changes.
\item \textbf{Semantic verification only where it is needed.} We combine sigmoid-based
  embedding matching with prompt ensembles and a conditional VLM judge invoked only inside an uncertainty interval, with contextual neighbour
  masking to isolate each candidate. This yields high semantic precision while keeping the VLM
  activation rate, and hence latency, low.
\item \textbf{An empirical study of ontological robustness.} On a real capture of
  hyper-realistic statues, our method removes the false positives that SAM 3 produces while
  preserving most of its recall, and a set of ablations quantifies the contribution of each
  component and the accuracy--latency trade-off of the uncertainty interval.
\end{itemize}

The code and models are publicly available.

\section{Related Work}

\paragraph{Promptable and universal segmentation.}
Instance and panoptic segmentation matured through closed-set detectors and mask heads such as
Faster R-CNN~\cite{fasterrcnn} and Mask R-CNN~\cite{maskrcnn}, and later through set-prediction
transformers such as DETR~\cite{detr}, DINO~\cite{dinodet}, Mask2Former~\cite{mask2former}, and
OneFormer~\cite{oneformer}, all of which segment a fixed vocabulary of classes. The Segment
Anything Model~\cite{sam} broke with this by making segmentation promptable from points, boxes,
and masks, trained on a billion-mask corpus, and spawned variants that improve mask
quality~\cite{hqsam}, efficiency~\cite{efficientsam}, granularity~\cite{semanticsam}, and prompt
diversity~\cite{seem}. SAM 2~\cite{sam2} extended the paradigm to video with a Hiera
backbone~\cite{hiera} and a streaming memory bank that conditions the current frame on past
predictions, enabling object tracking from a single annotation. SAM 3~\cite{sam3} adds Promptable
Concept Segmentation: given a noun phrase, it detects, segments, and tracks every matching
instance. ENEAS builds on this family for mask generation, using SAM 2.1 as its segmentation head,
but targets the reliability gap that remains once prompts become linguistic: identity under
absence and occlusion, and ontological correctness under visual ambiguity.

\paragraph{Open-vocabulary detection and segmentation.}
Transferring vision-language knowledge into detectors began with distillation from CLIP-like
models~\cite{vild} and image-level supervision~\cite{detic}, and continued with grounded
pre-training that treats detection as phrase grounding~\cite{glip}. Transformer detectors were
then made open-vocabulary at scale~\cite{owlvit,owlv2}, Grounding DINO~\cite{groundingdino} and its
successor~\cite{groundingdino15} fused language and vision features for open-set detection, and
YOLO-World~\cite{yoloworld} brought the capability to real-time regimes. On the segmentation side,
LSeg~\cite{lseg} aligned per-pixel embeddings with text and OpenSeg~\cite{openseg} aligned
caption words with predicted masks, ODISE~\cite{odise} exploited the representations of
text-to-image diffusion models, SAN~\cite{san} attached a light side network to a frozen CLIP, and
X-Decoder~\cite{xdecoder} and OpenSeeD~\cite{openseed} unified segmentation with other
vision-language or detection tasks in a single decoder. Grounded
SAM~\cite{groundedsam} chains an open-set detector with SAM to obtain text-conditioned masks.
These detect-then-segment systems are strong per-frame detectors but have no memory, which
produces the identity drift we analyse in Section~\ref{sec:tracking}. Florence~\cite{florence}
introduced a vision foundation model adaptable to detection, retrieval, and captioning, and
Florence-2~\cite{florence2} recast it as a sequence-to-sequence model that solves detection,
phrase grounding, and captioning through task prompts. Our method uses Florence-2 as
its region proposal component in both modes.

\paragraph{Referring and reasoning segmentation.}
Referring segmentation localises the single object described by an expression, in
images~\cite{lavt} and in video~\cite{referformer}, with recent benchmarks stressing motion
expressions~\cite{mevis}. A newer line couples multimodal large language models with mask decoders
so that the language model itself emits a segmentation token: LISA~\cite{lisa} and
GLaMM~\cite{glamm} for images, and VISA~\cite{visa}, VideoLISA~\cite{videolisa}, and
Sa2VA~\cite{sa2va} for video. These models put the language model in charge of answering the
prompt. SeC~\cite{sec}, on which our tracking mode builds, uses a large vision-language model
differently: not to answer the prompt, but to maintain a concept-level memory of the tracked
object once a spatial prompt has been given.

\paragraph{Video object segmentation with memory.}
Memory-based video object segmentation propagates a mask from an annotated frame through
space-time memory~\cite{stm}, long-term memory models~\cite{xmem}, object-level
readout~\cite{cutie}, and hierarchical propagation~\cite{deaot}. SAM 2 adopted the same principle,
and follow-ups extend its memory for long videos~\cite{sam2long}, make it motion-aware for
tracking~\cite{samurai}, or add distractor-aware memory~\cite{dam4sam}. DEVA~\cite{deva} and
SAM-Track~\cite{samtrack} decouple an image-level open-vocabulary segmenter from temporal
propagation, and multi-object trackers associate detections over time with~\cite{ovtrack} or
without~\cite{bytetrack} language. These systems either propagate a visual prompt through
appearance memory or fuse per-frame detections; none maintains a model of what the target is.
SeC~\cite{sec} augments SAM 2 with a vision-language model that
progressively builds a concept-level representation of the target from past keyframes and injects
it into the mask decoder when a scene change is detected. Our method provides the missing text
interface and inherits SeC's robustness to disappearance and scale change.

\paragraph{Vision-language embeddings.}
CLIP~\cite{clip} and ALIGN~\cite{align} established contrastive image-text pre-training with a
softmax objective over the pairs in a batch, and open reproductions at scale followed with
LAION-5B~\cite{laion5b}. SigLIP~\cite{siglip} replaced the softmax with a pairwise sigmoid loss,
so that each image-text pair is scored independently, and SigLIP 2~\cite{siglip2} improved
semantic understanding and localisation, adding a NaFlex variant that processes images at their
native aspect ratio in the spirit of NaViT~\cite{navit}. Prompt ensembles were introduced with
CLIP itself and later learned~\cite{coop}, and calibrating the resulting scores is a
long-standing concern~\cite{calibration}. Section~\ref{sec:evolution} shows why sigmoid
independence matters for filtering candidate crops.

\paragraph{Vision-language models as judges.}
Instruction-tuned vision-language models such as LLaVA~\cite{llava}, BLIP-2~\cite{blip2},
InternVL~\cite{internvl}, PaliGemma~\cite{paligemma}, and the Qwen-VL
series~\cite{qwen2vl,qwen25vl,qwen3vl} can answer open questions about an image. Language models
are increasingly used to judge the outputs of other models~\cite{llmjudge}, but vision-language
models are themselves prone to object hallucination~\cite{pope}, and eliciting explicit
reasoning~\cite{cot} improves accuracy at the cost of generating many additional tokens. Constrained
decoding~\cite{outlines} makes their verdicts machine-readable, and cascades that reserve
expensive models for hard inputs~\cite{frugalgpt}, or abstain when uncertain~\cite{selective},
trade accuracy against cost. ENEAS applies these ideas to segmentation: a small Qwen3-VL acts as a
conditional judge, restricted to the cases where embeddings alone are indecisive and answering
through a constrained schema without free-form reasoning.

\paragraph{Distractors in 3D reconstruction.}
Neural radiance fields~\cite{nerf} and 3D Gaussian Splatting~\cite{gs} assume a static scene, and
transient objects such as passers-by leave floaters and ghosts. NeRF in the Wild~\cite{nerfw}
models transients with per-image latents, RobustNeRF~\cite{robustnerf} and NeRF
On-the-go~\cite{nerfonthego} down-weight them with robust losses and uncertainty, and
SpotlessSplats~\cite{spotless} and WildGaussians~\cite{wildgaussians} bring the same ideas to
Gaussian splatting. A complementary line lifts 2D segmentation and language features into the 3D
representation~\cite{gaussiangrouping,saga,langsplat}. These methods handle distractors inside the reconstruction, from photometric residuals,
uncertainty, or generic visual features, without assuming what a distractor is. ENEAS instead
removes them semantically before reconstruction, with explicit masks and an explicit notion of what a
distractor is, which is what allows it to spare a statue while removing the visitor next to it.

\section{Method}
\label{sec:method}

ENEAS is a single method that takes a natural-language prompt and a set of frames and returns
binary masks. It is built on shared components: Florence-2-Large~\cite{florence2} grounds the
prompt into image regions, and a Segment Anything model produces the final masks. Depending on
whether the prompt refers to a specific instance (e.g., ``the blue painting'') or to a category
(e.g., ``person''), the method either propagates a single initialization through time or verifies every
candidate region semantically.

\paragraph{Problem statement.}
Given a collection of frames $\{\mathbf{I}_t\}_{t=1}^{N}$, $\mathbf{I}_t \in \mathbb{R}^{H \times W \times 3}$,
which may or may not be temporally ordered, and a natural-language prompt $p$, ENEAS returns
binary masks. When $p$ designates a specific instance, the output is one mask per frame,
$\mathbf{M}_t \in \{0,1\}^{H \times W}$, with $\mathbf{M}_t = \mathbf{0}$ whenever the object is not visible. When
$p$ names a category, the output is a set of masks per frame, $\{\mathbf{M}_t^{(i)}\}_{i=1}^{n_t}$, one per
instance, where the number of instances $n_t$ varies from frame to frame. Both cases share a
grounding operator $\mathcal{G}(\mathbf{I}, p)$ that returns image regions matching $p$, and a segmentation
operator $\mathcal{S}(\mathbf{I}, \mathbf{r})$ that turns a region $\mathbf{r}$ into a mask.

\subsection{Instance Tracking}
\label{sec:method-tracking}

For tracking a specific object defined by the user (e.g., ``the blue car''), ENEAS builds
upon the SeC architecture~\cite{sec}, which extends SAM 2 with a concept-level memory of the
target. Rather than matching appearance alone, the tracker maintains a high-level representation of
what the object is, which makes it robust to drastic viewpoint changes and to the target leaving
and re-entering the view. Since SeC natively supports only point-based interaction, we add a
text-driven initialization: on a reference frame $\mathbf{I}_{t_0}$, the grounding operator locates the
object described by the prompt, $\mathbf{r}_0 = \mathcal{G}(\mathbf{I}_{t_0}, p)$, and this single region initializes
the tracker. Every other mask is then obtained by propagation,
\begin{equation}
  \mathbf{M}_t = \mathcal{T}\big(\mathbf{I}_t \mid \mathbf{r}_0, \mathcal{H}_{t}\big),
\end{equation}
where $\mathcal{T}$ is the SeC tracker and $\mathcal{H}_{t}$ its memory of the frames already
processed, so that identity is enforced throughout: when the object is not visible, $\mathbf{M}_t$ is empty
rather than drifting to a look-alike. Two interaction modes are supported, points for maximum precision and natural language
for ease of use, and the reference frame may be chosen anywhere in the sequence.

\subsection{Semantic Discovery}
\label{sec:method-discovery}

\begin{figure*}[t]
  \centering
  \includegraphics[width=\linewidth]{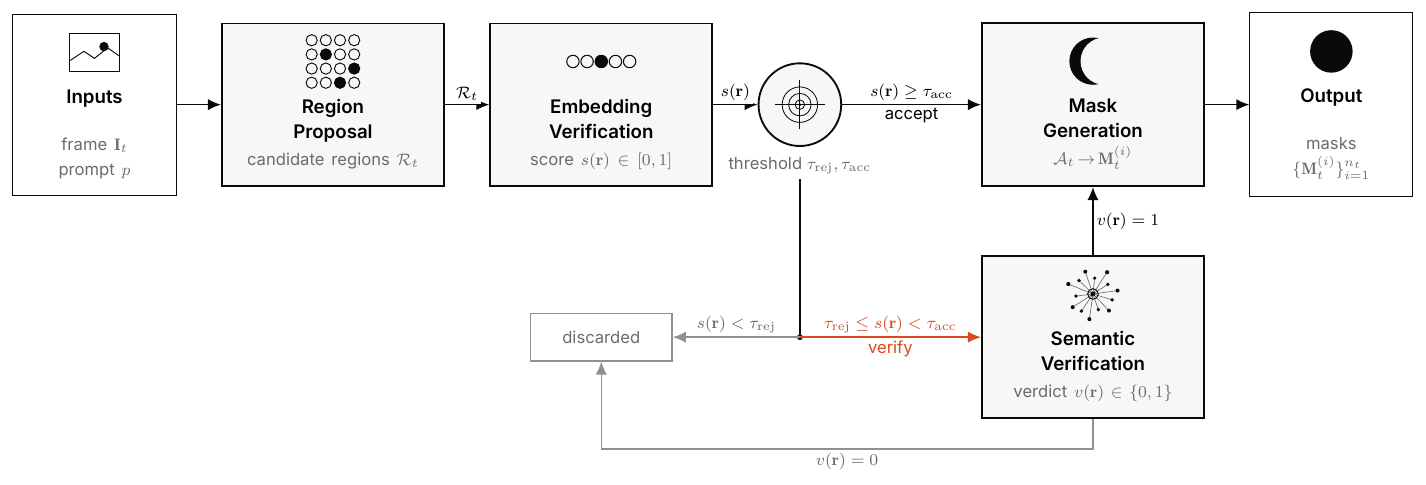}
  \caption{\textbf{ENEAS decision flow for semantic discovery.} For every frame $\mathbf{I}_t$ and prompt $p$, region proposal yields candidate regions $\mathcal{R}_t$; embedding verification scores each candidate with $s(\mathbf{r})$. Candidates above $\tau_{\mathrm{acc}}$ are accepted directly and those below $\tau_{\mathrm{rej}}$ are discarded; only the uncertainty interval in between is routed to semantic verification, where a vision-language model returns a verdict $v(\mathbf{r})$. The accepted set $\mathcal{A}_t$ is turned into one binary mask per instance by mask generation. Each stage is costlier than the previous one and sees fewer candidates.}
  \label{fig:fig3}
\end{figure*}

For semantic discovery, ENEAS re-evaluates every frame, so that new instances entering the
scene are found without re-prompting. It is organised as a cascade of stages of increasing cost,
each applied only to the candidates the previous one could not resolve. The complete decision flow
is visualized in Figure~\ref{fig:fig3}.

\paragraph{Region proposal.}
The grounding operator, instantiated with Florence-2~\cite{florence2}, proposes a set of candidate
regions for the requested category on every frame, $\mathcal{R}_t = \mathcal{G}(\mathbf{I}_t, p)$, and
duplicate proposals are merged. This stage is deliberately permissive: it
should miss nothing, at the cost of proposing distractors that later stages must remove.

\paragraph{Embedding verification.}
Each candidate is scored against the category with a vision-language embedding model,
SigLIP 2~\cite{siglip2}, that judges every image-text pair independently, so that the presence of
other objects in the crop does not suppress the score of the target. To make the scores robust to
the difference between the whole images the model was trained on and the crops it sees here,
several phrasings of the category are averaged into a single score $s(\mathbf{r}) \in [0,1]$ per candidate.
Two thresholds, $\tau_{\mathrm{rej}} < \tau_{\mathrm{acc}}$, split the candidates into three
groups:
\begin{equation}
  \mathbf{r} \;\mapsto\;
  \begin{cases}
    \text{accept}  & \text{if } s(\mathbf{r}) \geq \tau_{\mathrm{acc}},\\
    \text{discard} & \text{if } s(\mathbf{r}) < \tau_{\mathrm{rej}},\\
    \text{verify}  & \text{otherwise},
  \end{cases}
  \label{eq:decision}
\end{equation}
so that clearly matching candidates are accepted, clearly non-matching ones are discarded, and the
remainder, the \textit{uncertainty interval}, is deferred to the next stage. The two thresholds
are the knobs that trade latency against semantic rigor (Section~\ref{sec:thresholds}).

\paragraph{Semantic verification.}
Only candidates in the uncertainty interval are shown to a vision-language model,
Qwen3-VL~\cite{qwen3vl}, acting as a judge that returns a binary verdict $v(\mathbf{r}) \in \{0,1\}$. It is
asked whether the region truly is an instance of the category and not a look-alike such as a
statue, a mannequin, or a picture. Pixels belonging to
neighbouring candidates are masked out so that the judgement concerns the candidate alone, and the
model answers in a fixed format without free-form reasoning, which keeps the cost of each verdict
low. Because this stage is reached only by genuinely ambiguous candidates, its cost scales with the
ambiguity of the scene rather than with the number of objects in it.

\paragraph{Mask generation.}
The accepted set $\mathcal{A}_t$, made of the candidates accepted directly and those with
$v(\mathbf{r}) = 1$, is passed to the segmentation operator, instantiated with SAM 2~\cite{sam2} from the
Segment Anything family~\cite{sam}, giving one mask per instance, $\mathbf{M}_t^{(i)} = \mathcal{S}(\mathbf{I}_t, \mathbf{r}_i)$
for $\mathbf{r}_i \in \mathcal{A}_t$.

\section{Experiments \& Results}
\label{sec:experiments}

Before detailing the experimental results, we outline the evolution of our architecture. It is
important to note that while ENEAS addresses both instance tracking and semantic
discovery, the majority of our design analysis and ablation studies focus on semantic
discovery. This is where our method introduces its novel cascaded verification involving embeddings and
VLMs, which requires rigorous justification. Instance tracking, being a direct adaptation
of the SeC tracker, is evaluated primarily on its comparative robustness in temporal tracking
scenarios.

\subsection{Target Scenario and Datasets}
\label{sec:scenario}

ENEAS was designed with 3D reconstruction and novel view synthesis~\cite{nerf,gs} in mind. A capture for these
tasks is a collection of photographs or a video, often at low frame rates, whose views differ
widely in viewpoint, frequently lack a temporal order, and occlude the subject to varying degrees.
Before reconstruction, two questions must be answered for every view: what to keep and what to
remove. Instance tracking answers the first, isolating the object being reconstructed
across all views. Semantic discovery answers the second, removing every distractor of a class,
typically people, so that no ghost remains in the reconstructed scene. The costs of the two errors
are asymmetric: a false positive masks part of the asset and destroys it, whereas a false negative
leaves a visible artifact. This asymmetry is what motivates the precision-first design of
Section~\ref{sec:method}. The capture regime also makes the design affordable: most views contain
few or no distractors, so the embedding filter resolves them and the VLM is rarely invoked, and the
process is offline, so a few seconds per image amounts to minutes for a collection of hundreds of
views, negligible next to the reconstruction itself.

\textbf{Church Statues} is such a capture. It consists of frames extracted from a low frame rate
recording of a church interior, acquired for a 3D reconstruction, in which hyper-realistic
religious sculptures stand alongside visitors. The sculptures are what must be preserved and the
visitors what must be removed, so the capture naturally maximizes ontological ambiguity for the
prompt ``person''. Ground-truth instance annotations were produced manually by the authors. The
remaining sequences, \textbf{Moving Boxes} (an indoor scene in which people carry and stack boxes
among furniture, occluding each other and the chairs) and \textbf{Blue Painting} (a dynamic indoor
sequence in which a single painting undergoes viewpoint changes, occlusions, and close-ups),
together with the external \textbf{SA-Co/VEval} benchmark, are conventional video and are used to verify that the robustness obtained for the capture scenario
holds beyond it.

\subsection{Design Evolution and Justification}
\label{sec:evolution}

The architecture proposed in this work is the result of a rigorous iterative process aimed at
resolving specific trade-offs between \textbf{recall}, \textbf{precision}, and computational
latency. In the context
of this study, we define recall as the system's capacity to detect every instance of the target
category, thereby minimizing false negatives. Conversely, precision denotes the ability to
rigorously distinguish the target concept from semantic distractors, ensuring that false positives
are eliminated. In this section, we detail the evolution of our method through four distinct
development phases, analyzing the theoretical limitations of prior approaches that necessitated our
final design.

\paragraph{Phase 1: Limitations of Exhaustive Scene Parsing and Composite Visual Prompting.}
Our initial approach employed a detect-filter-validate strategy designed to maximize semantic
understanding. We implemented an exhaustive dense object captioning mechanism aiming to generate
bounding boxes accompanied by rich textual descriptions for all salient elements within the scene.
Subsequently, to alleviate the load on the validation stage, we applied an intermediate textual
filter, involving the computation of semantic similarity between the generated descriptions and the
target prompt using text-only embeddings.

For the final validation, we attempted to optimize throughput using a composite visual prompting
strategy. Instead of processing crops individually, we constructed a single query image containing
the full frame annotated with color-coded, numbered bounding boxes, supplemented by a grid of
spatially clustered crops to preserve resolution. The downstream vision-language model was queried
with a complex, multi-step prompt instructed to iterate through and validate each numbered instance
sequentially.

While theoretically efficient, this approach exhibited critical failure modes inherent to the dense
description paradigm. First, the textual filter was strictly dependent on the generative fidelity of
the region proposal network, suffering from \textit{Open-Set Hallucinations}. For instance, the generation
module would frequently assign arbitrary proper nouns (e.g., labeling a statue as ``Saint
Christian'') to generic objects, effectively bypassing the semantic filter. Furthermore, this
approach suffered from spatial aggregation, frequently merging distinct semantic entities, such as a
chair and a surrounding table, into single bounding boxes. This rendered precise instance
segmentation impossible. Ultimately, the high cognitive load of the multi-target association task
within a single prompt necessitated the activation of deep reasoning capabilities to prevent
attention drift. This dependency introduced prohibitive latency, starting at a minimum of 15 seconds
per frame and escalating significantly in dense scenes. Consequently, this approach proved inviable
for video processing.

\paragraph{Phase 2: The Softmax Bottleneck in Contrastive Filtering.}
To address the spatial aggregation and recall issues encountered in the previous phase, we
transitioned to a text-conditioned region proposal strategy. By explicitly prompting the detector
with the target category, we improved instance separation but introduced a high rate of false
positives. We attempted to filter these candidates using contrastive vision-language embedding
matching. Initially, we computed the softmax probability of the target text against generic negative
classes such as ``background'' and ``other object.'' To improve granularity, we implemented a
secondary object detection pass to identify specific objects present in the scene (e.g.,
``microphone'', ``table'') and included these dynamic labels in the softmax computation.

However, this approach revealed the fundamental deficiency of applying global softmax normalization
to this task. The mutual exclusivity constraint inherent in the softmax function forces class
probabilities to compete. In scenarios where a target object co-occurred with highly salient
attributes identified in the secondary pass, such as a person holding a microphone, the model
suppressed the target class probability in favor of the accessory's label. This phenomenon resulted
in score suppression that made it impossible to define a stable acceptance threshold.

\paragraph{Phase 3: Sigmoidal Independence and Aspect Ratio Optimization.}
To resolve the score suppression issue, we implemented a sigmoid-based vision-language embedding
matching. Crucially, this architecture employs a pairwise sigmoid loss, treating each image-text
pair as an independent binary classification problem. This formulation effectively decouples the
target object's score from the presence of other semantic concepts in the crop. We specifically
utilized an encoder optimized for native aspect ratios, which preserves the dense features of our
variable-size crops better than standard fixed-resolution encoders.

Despite these architectural improvements, we observed a performance degradation caused by the domain
shift between the model's full-image pre-training and our crop-level inference inputs. To mitigate
this, we implemented a prompt ensemble strategy, averaging embeddings across multiple template
variations to enforce robustness in the visual signal.

\paragraph{Phase 4: Optimization of the Semantic Judge.}
Despite the improved embedding filter, a \textit{zone of uncertainty} persisted where visual embeddings
alone failed to resolve ontological ambiguities, such as hyper-realistic statues. To address this,
we integrated a lightweight VLM. Initially, executing the model with its native deep reasoning
capability proved accurate but incurred significant latency due to extensive token generation.
Furthermore, the model struggled with contextual interference from neighboring objects within the
crop.

To mitigate this, we apply black masking to isolate the target pixel-wise. Subsequently, we disabled
the deep reasoning mechanism in favor of a structured analytic prompt with explicit critical
thinking constraints. This approach forces the model to verify the ontology without the overhead of
generating internal reasoning tokens. By restricting this validation to the uncertainty interval, we
achieve high semantic precision while reducing verification latency to approximately one second on
resource-constrained hardware.

\subsection{Hardware and Evaluation Setup}
All efficiency and latency experiments reported in this study were conducted in a standardized cloud
inference environment to ensure reproducibility. The hardware configuration consisted of a single
NVIDIA L4 GPU based on the Ada Lovelace architecture with 24 GB of VRAM, paired with an Intel Xeon
CPU running 12 virtual cores at 2.20 GHz. This setup was selected to simulate a realistic,
cost-effective production scenario rather than a high-end high-performance computing environment. It
is important to note that the reported latency figures encompass the full end-to-end pipeline,
including data loading from storage and CPU-bound preprocessing steps such as normalization and
resizing, which constitutes a non-negligible portion of the execution time in virtualized cloud
environments.

\subsection{Robustness in Instance Tracking}
\label{sec:tracking}

We evaluate the efficacy of ENEAS in handling instance disappearance, occlusion, and scale
variation, which are critical failure modes for prior text-guided trackers. We conducted a
qualitative analysis on the Blue Painting sequence, targeting a specific object defined as ``blue
painting'' that undergoes severe viewpoint changes and partial occlusions.

\begin{figure*}[t]
  \centering
  \includegraphics[width=\linewidth]{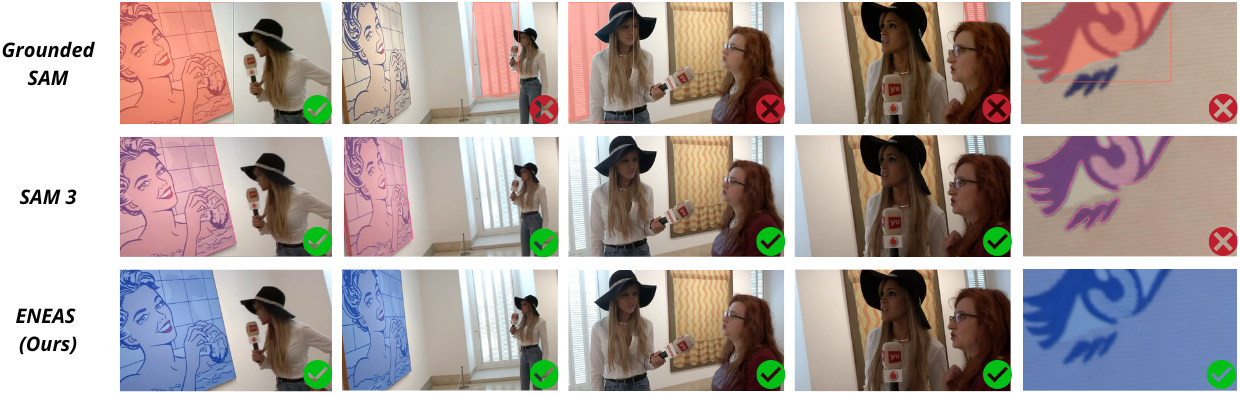}
  \caption{\textbf{Temporal Robustness and Spatial Integrity Comparison.} The target is the ``blue painting''. Row 1 (Grounded SAM): Suffers from severe drift when the target is occluded or absent, incorrectly segmenting curtains and hair (red crosses). It also fails to capture the full object during close-ups (last column). Row 2 (SAM 3): Improves temporal stability but suffers from spatial fragmentation in the close-up, segmenting internal details instead of the full object instance. Row 3 (ENEAS): correctly handles target absence by outputting no mask and maintains spatial coherence, capturing the entire object even under extreme scale changes.}
  \label{fig:fig4}
\end{figure*}

\paragraph{Robustness to Target Absence.}
As illustrated in Figure~\ref{fig:fig4} (Row 1), baselines relying on per-frame detection such as Grounded
SAM~\cite{groundedsam} (using Grounding DINO~\cite{groundingdino} and SAM 2.1~\cite{sam2}) exhibit
significant identity drift. When the target object exits the field of view
or is heavily occluded, the model hallucinates positive detections on background elements such as
curtains or distractor objects like the subject's hair to satisfy the text prompt. In contrast,
ENEAS (Row 3) leverages the propagated temporal memory of the SeC architecture to enforce
identity constraints. It correctly reports a \textbf{True Negative} (zero mask) when the specific instance is
not visible. This capability prevents the accumulation of false positives in long-term tracking.

\paragraph{Spatial Integrity under Scale Change.}
We further observed limitations in the spatial coherence of foundational models during extreme
close-ups (Figure~\ref{fig:fig4}, Column 5). Both Grounded SAM and SAM 3~\cite{sam3} suffer from \textit{spatial fragmentation} as
they segment only high-frequency details such as specific brushstrokes or figures within the
painting rather than the object as a holistic entity. Our approach maintains the semantic integrity
of the ``painting'' object and provides a complete mask even when the object occupies the entire
view. This demonstrates that our approach effectively bridges the gap between high-level semantic
prompts and low-level temporal consistency.

\subsection{Discovery of New Instances}

We evaluate the ability of the system to continuously discover and segment multiple instances of a
category in cluttered environments. This capability is essential for real-world video analysis. To
this end, we utilize two sequences: Church Statues targeting ``real person'' and Moving Boxes
targeting ``chair,'' which features high geometric clutter due to stacked boxes and furniture.

\begin{figure*}[t]
  \centering
  \includegraphics[width=0.95\linewidth]{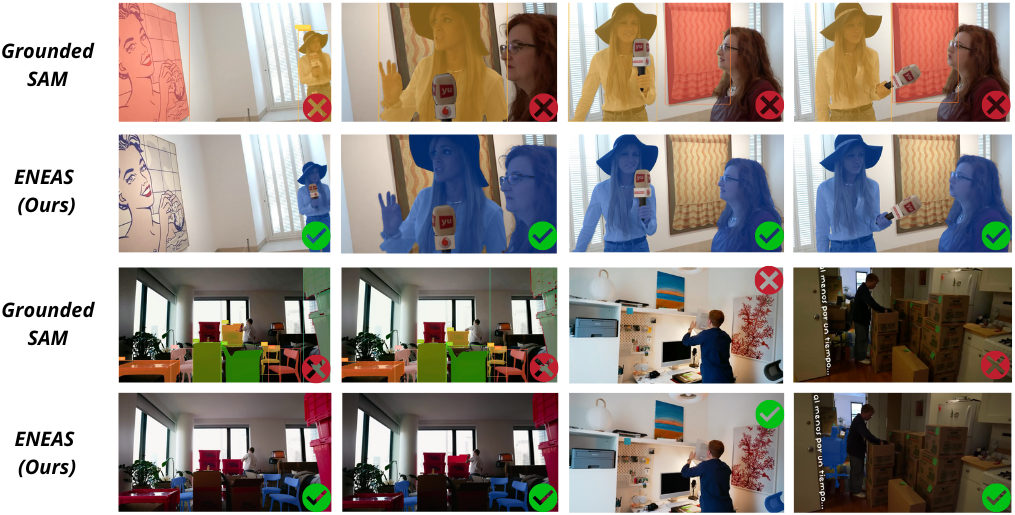}
  \caption{\textbf{Continuous Discovery and Robustness to Clutter.} Comparison of Grounded SAM versus ENEAS in multi-instance discovery. Top Rows (Prompt: ``real person''): Grounded SAM hallucinates masks on the background curtains and fails to maintain consistent IDs. ENEAS correctly segments only the valid subjects throughout the sequence. Bottom Rows (Prompt: ``chair''): In a highly cluttered moving scene, Grounded SAM suffers from catastrophic over-segmentation by labeling boxes and tables as chairs. ENEAS exhibits superior semantic filtering, exclusively segmenting the target objects while ignoring significant geometric clutter.}
  \label{fig:fig5}
\end{figure*}

As demonstrated in Figure~\ref{fig:fig5}, baseline methods such as Grounded SAM~\cite{groundedsam} (Rows 1 and 3) exhibit severe
semantic drift in multi-instance scenarios. In Moving Boxes (Row 3), when prompted for
``chair,'' the baseline fails to distinguish between the target object and the surrounding clutter.
It incorrectly segments tables, cardboard boxes, and walls as positive instances. This
over-segmentation renders the output unusable for automated inventory or analysis. Conversely,
ENEAS (Row 4) demonstrates surgical precision. By leveraging VLM verification, it
successfully identifies valid chairs (marked in blue) while
consistently rejecting geometrically similar distractors such as stacked boxes, even under partial
occlusion.

Furthermore, unlike propagation-only models such as SAM 2.1 or its grounded variant, which are
limited to instances annotated in the initial frame, our method continuously re-evaluates the scene.
This approach allows for the seamless detection of new instances entering the field of view, such as
a new person walking in or a chair being revealed, without requiring user re-prompting. This
effectively solves the static initialization bottleneck inherent in previous state-of-the-art
trackers.

\subsection{Semantic Rigor: Direct Comparison to SAM 3}

We benchmark our system directly against SAM 3~\cite{sam3}, the newly released state-of-the-art foundation model
for Promptable Concept Segmentation. While SAM 3 represents a significant leap in open-vocabulary
localization, we evaluate its performance in scenarios dominated by ontological ambiguity, where the
challenge lies not in detecting a shape but in verifying its essence, such as distinguishing real
humans from artistic representations.

\begin{figure*}[t]
  \centering
  \includegraphics[width=\linewidth]{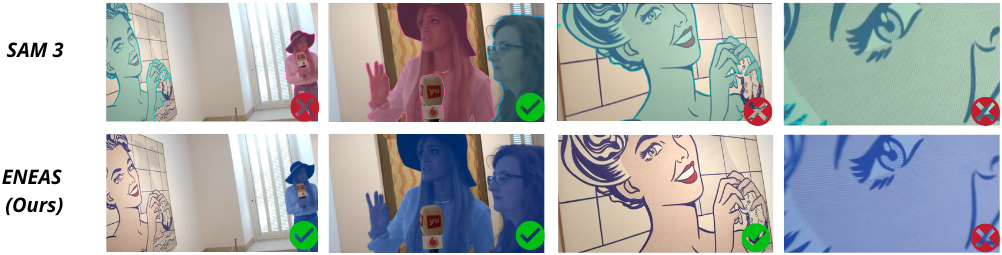}
  \caption{\textbf{Semantic Ambiguity Challenge I -- Artistic Representations.} The prompt is ``real person''. Row 1 (SAM 3): Lacks ontological filtering, repeatedly segmenting the large pop-art painting as a person due to visual similarity. Row 2 (ENEAS): leverages VLM verification to correctly identify the painting as a non-living representation, focusing only on real human subjects. Note that both models struggle in the extreme close-up (last column) due to the lack of surrounding context, but ENEAS minimizes false positives across the full sequence.}
  \label{fig:fig6}
\end{figure*}

\begin{figure*}[t]
  \centering
  \includegraphics[width=\linewidth]{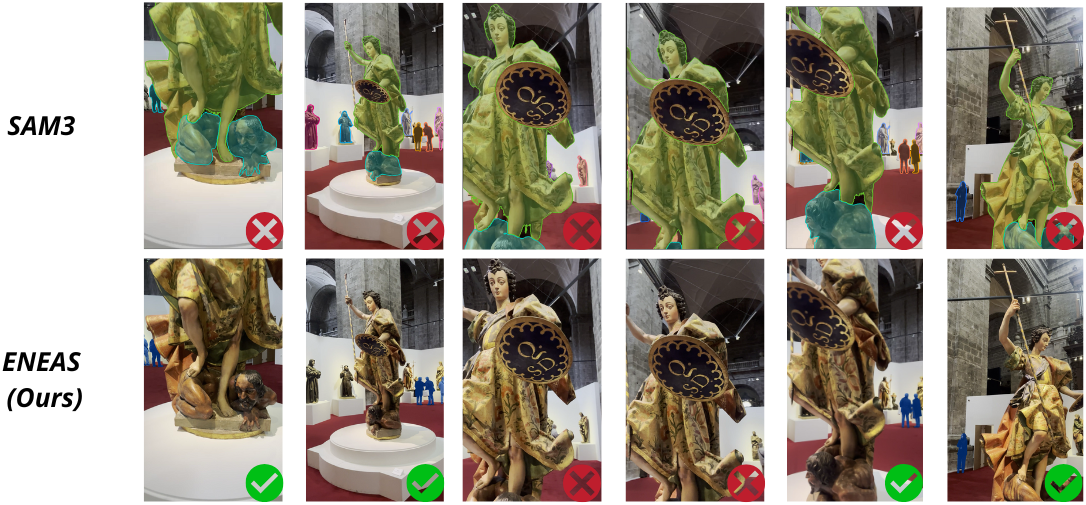}
  \caption{\textbf{Semantic Ambiguity Challenge II -- Hyper-Realistic Statues.} The prompt is ``person''. Row 1 (SAM 3): Fails to distinguish between statues and humans, generating masks for all sculptures. Row 2 (ENEAS): Demonstrates semantic rigor by rejecting the statues. Note that both models exhibit minor False Negatives (red crosses) on background pedestrians, but ENEAS eliminates the massive False Positive noise present in the SOTA baseline.}
  \label{fig:fig7}
\end{figure*}

\paragraph{Qualitative Analysis.}
Figures~\ref{fig:fig6} and~\ref{fig:fig7} illustrate two distinct failure modes inherent to visual-only foundation
models. In the Pop-Art Test depicted in Figure~\ref{fig:fig6}, a large-scale painting dominates the background.
Despite the clear semantic distinction between a flat artwork and a three-dimensional human, SAM 3
consistently segments the painted figure as a valid person driven by the strong visual similarity of
the facial features. Similarly, in the Statue Test shown in Figure~\ref{fig:fig7}, SAM 3 fails to distinguish
between hyper-realistic religious sculptures and living humans. The model generates high-confidence
masks for nearly every statue in the scene.

In contrast, ENEAS demonstrates superior semantic reasoning in both scenarios. By
routing ambiguous crops to the vision-language model referee, the system successfully rejects both
the two-dimensional painting and the three-dimensional sculptures in the majority of frames,
focusing exclusively on the living subjects. It is worth noting that in extreme close-ups where
context is entirely absent (e.g., Figure~\ref{fig:fig6}, Column 4), our method can exhibit similar leakage to the
baseline due to the lack of surrounding visual cues for the VLM, but crucially, it significantly
reduces the frequency and severity of these false positives across the sequence compared to SAM 3.
This validates that our approach effectively bridges the gap between visual perception and cognitive
understanding.

\paragraph{Quantitative Benchmark.}
To quantify this performance gap, we evaluated all models on the full Church Statues capture.
This benchmark rigorously tests the ability to distinguish between living subjects and realistic
artifacts under the prompt ``person''. The results, summarized in Table~\ref{tab:foundation-comparison}, reveal a
stark contrast between visual foundation models and our proposed architecture.

Unlike tracking-oriented benchmarks, this setting is designed to measure \emph{ontological
correctness} (i.e., whether each predicted instance truly belongs to the target concept). For this
reason, we report \textbf{Precision/Recall/F1} instead of HOTA-style metrics in this subsection.
\textbf{Precision} captures semantic purity (how many predicted instances are truly valid),
\textbf{Recall} captures recovery of valid targets, and \textbf{F1} summarizes their balance. The
complementary tracking metrics (\textbf{HOTA/DetA/AssA/LocA/TETA}) are reported separately in the
SA-Co/VEval analysis below.

\begin{table*}[t]
  \centering
  \caption{Comparison against Foundation Models on Church Statues}
  \begin{tabular}{lccc}
    \toprule
    \thd{Model} & \thd{Precision (\%)} & \thd{Recall (\%)} & \thd{F1-Score (\%)} \\
    \midrule
    SAM 3 & 11.1 & \best{83.7} & 19.5 \\
    ENEAS-2B (Ours) & \runnerup{94.7} & 73.5 & \runnerup{82.8} \\
    ENEAS-4B (Ours) & \best{97.5} & \runnerup{79.6} & \best{87.6} \\
    \bottomrule
  \end{tabular}
  \label{tab:foundation-comparison}
\end{table*}

As detailed in Table~\ref{tab:foundation-comparison}, SAM 3 achieves the highest recall, confirming its status as an exceptionally
powerful detector capable of identifying instances that other visual backbones miss. However, this
sensitivity comes with a catastrophic False Positive rate, resulting in a precision of just 11.1\%.
In practical terms, for every correct detection of a person, SAM 3 hallucinates approximately eight
artifacts.

Our default configuration (ENEAS-2B) accepts a trade-off in recall to enforce a strict precision of
94.7\%. The resulting F1-Score represents a four-fold improvement over the state-of-the-art
baseline. Furthermore, scaling the semantic judge to the 4B parameter variant yields a substantial
performance boost. By recovering valid instances that were ambiguous to the smaller model, the 4B
variant improves recall while maintaining near-perfect semantic purity. The resulting F1-Score of
87.6\% confirms that our architecture scales effectively with model capacity, offering a
configurable trade-off where the 2B model serves as a highly efficient baseline and the 4B model
provides maximum rigor for offline processing.

\paragraph{Discussion: Visual Priors versus Ontological Reasoning.}
The performance gap observed in the quantitative results and qualitative figures highlights a
fundamental limitation in current foundation models like SAM 3, which is the dominance of
\textit{visual priors} over \textit{ontological reasoning}. Having been trained on massive datasets to maximize mask quality
and recall, SAM 3 develops strong priors for structural coherence. Consequently, when prompted with
a concept such as ``person'', the model activates on high-level features including facial symmetry,
limb structure, and skin-like textures. This reliance causes hyper-realistic statues or large-scale
portraits, which perfectly satisfy these structural constraints, to become indistinguishable from
real humans within the model's latent space. Essentially, the model effectively solves the
segmentation task by identifying the object's boundary, yet fails the grounding task by
misidentifying the nature of the object.

Our method addresses this deficiency by structurally decoupling geometric proposal from
semantic verification. In this framework, the embedding filter serves as a fast visual gate,
handling the coarse rejection of non-conforming shapes and background elements. Crucially, the
integration of a VLM introduces a necessary cognitive layer that transcends simple visual matching.
Unlike segmentation models that process local texture cues, the vision-language model analyzes the
global context and material properties of the crop, such as identifying the flatness of a pop-art
painting or the granular texture of a stone statue, to make a high-level ontological judgment
regarding the subject's animacy. This design ensures that the system does not merely locate
shapes that visually resemble the target category but successfully verifies entities that
ontologically belong to it. By explicitly enforcing this reasoning step, our approach effectively
bridges the semantic gap where \textit{visual-only} foundation models inevitably drift.

\paragraph{SA-Co/VEval Benchmark (SA-V Test).}
In addition to Church Statues, we evaluate ENEAS and SAM 3 on an SA-V Test
subsample from SA-Co/VEval. We adopt a prompt-based split: samples whose prompt refers to a
specific object instance (e.g., ``a gold watch'') are assigned to \textit{Instance Tracking}, while
prompts that may match multiple objects are assigned to \textit{Semantic Discovery}. Both methods
are evaluated with the official SAM 3 SA-Co/VEval evaluator under identical settings.

We report HOTA together with its standard decomposition (DetA, AssA, LocA), as well as
TETA; all metrics are \textit{higher-is-better}. HOTA
summarizes overall tracking quality, DetA captures detection quality, AssA captures temporal
association consistency, and LocA measures localization quality~\cite{hota,teta}.
Results are reported in Tables~\ref{tab:sa_co_unique} and~\ref{tab:sa_co_generic}.

\begin{table*}[t]
\centering
\noindent
\begin{minipage}[t]{0.49\linewidth}
  \centering
  \captionof{table}{SA-Co/VEval --- Instance Tracking}
  \label{tab:sa_co_unique}
  \begin{tabular}{lcc}
    \toprule
    \thd{Metric} & \thd{ENEAS} & \thd{SAM 3} \\
    \midrule
    HOTA (\%) & \textbf{26.70} & 26.51 \\
    DetA (\%) & \textbf{7.92} & 7.84 \\
    AssA (\%) & \textbf{90.77} & 90.18 \\
    LocA (\%) & 87.86 & \textbf{89.22} \\
    TETA (\%) & \textbf{17.86} & 16.65 \\
    \bottomrule
  \end{tabular}
\end{minipage}\hfill
\begin{minipage}[t]{0.49\linewidth}
  \centering
  \captionof{table}{SA-Co/VEval --- Semantic Discovery}
  \label{tab:sa_co_generic}
  \begin{tabular}{lcc}
    \toprule
    \thd{Metric} & \thd{ENEAS} & \thd{SAM 3} \\
    \midrule
    HOTA (\%) & \textbf{9.23} & 9.19 \\
    DetA (\%) & 6.16 & \textbf{6.21} \\
    AssA (\%) & \textbf{14.28} & 13.85 \\
    LocA (\%) & 74.25 & \textbf{75.38} \\
    TETA (\%) & \textbf{9.53} & 9.10 \\
    \bottomrule
  \end{tabular}
\end{minipage}
\end{table*}

\par\vspace{0.9em}

Overall, the comparison reveals a consistent trade-off across both splits. ENEAS improves end-to-end
tracking metrics (HOTA and TETA) primarily through stronger association quality
(AssA), with comparable detection quality (DetA), whereas SAM 3 remains slightly
better in localization (LocA). Since HOTA jointly reflects detection, association, and
localization, the gains in AssA are
sufficient for our method to obtain a better overall tracking profile. This behavior is especially
relevant for language-conditioned video segmentation, where \textit{temporal identity consistency} is often
the dominant failure mode.

\section{Ablation Study}

To validate our architectural decisions, we conducted
a series of rigorous ablation studies focusing on \textbf{component necessity}, \textbf{threshold sensitivity}, and
computational efficiency. To ensure these tests meaningfully challenge the semantic reasoning
capabilities of the system, all experiments were performed on Church Statues. This
capture comes from a real acquisition for 3D reconstruction (Section~\ref{sec:scenario}) and naturally
maximizes ontological ambiguity, containing high densities of hyper-realistic sculptures alongside
real human subjects. We posit that standard benchmarks such as
COCO~\cite{coco}, LVIS~\cite{lvis}, or the video object segmentation suites DAVIS~\cite{davis},
YouTube-VOS~\cite{ytvos}, MOSE~\cite{mose}, and OVIS~\cite{ovis} lack the semantic complexity required to isolate the contribution of the VLM referee.
Consequently, this domain serves as the optimal testbed for evaluating the robustness of
our proposed architecture.

\subsection{Component Necessity Analysis}

To demonstrate the necessity of the VLM as a semantic referee, we evaluated distinct configurations
of our method on Church Statues. The quantitative results are summarized in
Table~\ref{tab:semantic-verification}.

\begin{table*}[t]
  \centering
  \caption{Impact of Semantic Verification Modules on Church Statues}
  \label{tab:semantic-verification}
  \small
  \resizebox{\linewidth}{!}{%
  \begin{tabular}{llccc l}
    \toprule
    \thd{Method Configuration} & \thd{Verification Logic} & \thd{Precision (\%)} & \thd{Recall (\%)} & \thd{F1-Score (\%)} & \thd{Behavior} \\
    \midrule
    RPN (Baseline) & Text Prompt Only & 10.5 & \best{79.6} & 18.6 & Severe hallucination of artifacts \\
    RPN + Embeddings & Flex Thresh.\ (T $>$ 0.50) & 55.7 & \best{79.6} & 65.5 & Filters noise, keeps realistic statues \\
    RPN + Embeddings & Strict Thresh.\ (T $>$ 0.80) & 94.3 & 67.4 & 78.6 & Misses valid instances \\
    ENEAS-2B (Ours) & Adaptive (0.10 $<$ T $<$ 0.90) & \runnerup{94.7} & 73.5 & \runnerup{82.8} & Optimal balance (speed vs performance) \\
    ENEAS-4B (Ours) & Adaptive (0.10 $<$ T $<$ 0.90) & \best{97.5} & \best{79.6} & \best{87.6} & Best performance, slower \\
    \bottomrule
  \end{tabular}
  }
  \vspace{0.4em}
  \caption*{\footnotesize \textit{Note:} RPN = Region Proposal Network (Florence-2); Embeddings = Embedding Matching (SigLIP); ENEAS includes the VLM Referee.}
\end{table*}

As observed, the baseline utilizing only a region proposal network achieves the highest theoretical
recall but suffers from \textbf{catastrophic precision}, rendering it unusable for uncurated environments due
to its inability to distinguish ontological differences. Integrating embedding matching as a visual
filter introduces a critical trade-off: a permissive threshold fails to filter hyper-realistic
statues, while a strict threshold successfully removes artifacts but aggressively discards valid
instances in challenging lighting or occlusion.

Our approach resolves this dilemma. By utilizing the 2B VLM solely for the
uncertainty interval, we recover True Positives rejected by the strict embedding matching while
maintaining a minimal False Positive rate. Furthermore, deploying the larger 4B model variant fully
recovers the recall to the baseline level while achieving near-perfect precision, yielding the
highest overall F1-Score. However, as discussed in Section~\ref{sec:vlm-size}, this performance gain comes with a
significant latency penalty.

It is important to note that the system's recall is inherently bounded by the initial region
proposal network. The False Negatives observed in the baseline persist across all configurations
because the detector failed to generate candidate bounding boxes for these instances. Qualitative
analysis reveals that these missed instances were primarily located in the background or heavily
occluded, highlighting the proposal bottleneck inherent in two-stage architectures. Our contribution
focuses on maximizing the precision of the detected proposals, ensuring that the final output is
semantically reliable.

\subsection{Threshold Sensitivity and Trade-off Analysis}
\label{sec:thresholds}

The acceptance and rejection thresholds for the visual embedding filter are configurable
parameters that allow adaptation to different latency and precision targets. We analyzed their
sensitivity using the 2B model variant on Church Statues to identify a practical ``Robust
Mode'' preset.

\begin{table*}[t]
  \centering
  \caption{Impact of Uncertainty Interval on Accuracy and Latency}
  \small
  \resizebox{\linewidth}{!}{%
  \begin{tabular}{lcccc}
    \toprule
    \thd{Configuration Profile} & \thd{Interval} & \thd{VLM Activation Rate (\%)} & \thd{Avg.\ Latency (s)} & \thd{F1-Score (\%)} \\
    \midrule
    Aggressive / Fast & 0.40--0.60 & 16.2 & \textbf{1.04} & 74.3 \\
    Balanced & 0.15--0.75 & 59.4 & 2.48 & 79.6 \\
    Robust & 0.10--0.90 & 78.1 & 3.29 & \textbf{82.8} \\
    VLM-Only & 0.00--1.00 & 100 & 5.05 & 79.1 \\
    \bottomrule
  \end{tabular}
  }
  \label{tab:uncertainty-interval}
\end{table*}

As detailed in Table~\ref{tab:uncertainty-interval}, the choice of uncertainty interval dictates the operational profile of the
system, creating a distinct trade-off between semantic purity and instance recovery. The Aggressive
configuration offers the lowest latency by resolving the majority of cases via high-speed visual
embeddings. However, this efficiency comes at the cost of significant semantic leakage, admitting
false positives that yield mid-range visual confidence scores. Expanding the interval to a Balanced
configuration effectively maximizes recall, successfully recovering instances that are visually
ambiguous, but at the cost of admitting high-confidence artifacts and increasing latency.

Our recommended \textbf{Robust} configuration (0.10--0.90) prioritizes semantic reliability and represents
the Pareto-optimal solution for ontologically ambiguous domains. By enforcing a strict acceptance threshold, it
successfully rejects nearly all artifacts, achieving the highest overall F1-Score. Crucially, this
hybrid approach reduces latency by approximately 35\% compared to the VLM-Only baseline. The
degradation in accuracy observed in the VLM-Only setting (F1 dropping from 82.8\% to 79.1\%)
suggests a phenomenon of over-reasoning: by forcing the 2B model to adjudicate unambiguous samples
(often small or heavily occluded crops where visual cues are sparse) the system introduces
stochastic errors that the simpler, robust visual embedding filter would have correctly handled. For
applications where this trade-off is unacceptable and maximum verification rigor is required
regardless of latency, upgrading to the 4B model variant (as discussed in Section~\ref{sec:vlm-size}) mitigates
these stochastic errors, albeit at a higher computational cost. This confirms that reserving the 2B
VLM solely for genuinely ambiguous semantic cases maximizes both efficiency and global accuracy in
the default configuration.

\subsection{Scenario Adaptability and Efficiency}

A key advantage of ENEAS is its adaptability to scene complexity. While the
Robust Configuration is essential in ontologically ambiguous scenes such as Church Statues to
filter semantic decoys, standard
scenarios allow for relaxed thresholds to maximize throughput. To quantify this efficiency gain, we
evaluated the system on the Moving Boxes sequence. This scenario represents a realistic indoor environment
which, while ontologically standard, presents visual challenges such as dynamic occlusions where
subjects are partially hidden by large boxes.

For this experiment, we utilized a \textbf{Fast Mode} configuration by lowering the acceptance threshold to
0.40. Under these conditions, the system achieved a latency of 1.14 seconds per frame, representing
a \textbf{three-fold speedup} compared to the robust execution on Church Statues (3.29 seconds).
This efficiency gain stems from the reduction in vision-language model activation rate, which
dropped from 78.1\% on Church Statues to 27.0\% on Moving Boxes. Crucially, despite the
relaxed threshold and the presence of occlusions, the visual embeddings successfully resolved the
majority of instances, maintaining a near-perfect F1-Score of 98.0\% with zero False Positives. This
confirms that the method is content-aware, dynamically allocating computational resources only
when ontological ambiguity is strictly present, and behaving as a lightweight filter in standard
processing workflows.

\subsection{VLM Size Analysis}
\label{sec:vlm-size}

Finally, we assessed the impact of the vision-language model capacity on system latency and
accuracy. We compared the default 2B model against the larger 4B variant on Church Statues
using the robust configuration.

\begin{table*}[t]
  \centering
  \caption{Impact of VLM Model Size on Efficiency and Accuracy}
  \begin{tabular}{lccc}
    \toprule
    \thd{Model variant} & \thd{Avg.\ Latency (s)} & \thd{F1-Score (\%)} & \thd{Speed Factor} \\
    \midrule
    2B & \textbf{3.29} & 82.8 & 1.0x (Baseline) \\
    4B & 5.02 & \textbf{87.6} & 1.5x Slower \\
    \bottomrule
  \end{tabular}
  \label{tab:vlm-size}
\end{table*}

As illustrated in Table~\ref{tab:vlm-size}, the 4B model acts as a semantic upper bound, achieving a remarkable
F1-Score of 87.6\% by successfully resolving the remaining ambiguous cases that challenged the
smaller model. However, this gain comes with a severe computational penalty: latency increases by
52\%, reaching 5.02 seconds per frame.

Consequently, we establish the \textbf{2B model as the default configuration} of ENEAS, as it
represents the optimal efficiency-accuracy trade-off for general video processing. However, since
our architecture is modular, we expose the \textbf{4B model} as a configurable option for offline
applications where maximum semantic rigor takes precedence over runtime latency. This flexibility
ensures the framework addresses both real-time constraints and high-fidelity segmentation needs.

\section{Conclusion}

We presented ENEAS, a unified method for text-guided segmentation that handles both specific
instance tracking and semantic discovery over ordered video and unordered image collections.
For instance tracking, a text-driven initialization stage turns the memory-based SeC tracker into
a text-prompted one, and the resulting system reports target absence instead of drifting and keeps
the target whole under extreme scale changes. For semantic discovery, the method decouples geometric
proposal from semantic verification: an open-vocabulary region proposal, a sigmoid-based
embedding filter with prompt ensembles, and a conditional VLM judge that is consulted only inside
an uncertainty interval. On Church Statues, a real capture of hyper-realistic sculptures, this design raises
F1 from 19.5\% for SAM 3 to 82.8\% with a 2B judge and 87.6\% with a 4B judge, while keeping
precision above 94\%; on SA-Co/VEval it matches or improves the overall tracking profile of SAM 3.
The ablations show that the embedding filter alone cannot be thresholded into both high precision
and high recall, that the VLM is necessary precisely in the interval where embeddings are
indecisive, and that consulting it everywhere is both slower and slightly less accurate than
consulting it selectively.

\paragraph{Limitations.}
The recall of semantic discovery is bounded by the region proposal stage: instances that
Florence-2 never proposes, typically small, distant, or heavily occluded ones, cannot be recovered
downstream. The VLM judge relies on context; in extreme close-ups where the crop contains no cues
about material or scene, it can accept the same artifacts as visual-only models; the same holds
for small or low-resolution crops, such as distant objects, where the judge has little to judge.
The cost of verification grows with the number of candidates per view, so dense scenes such as
crowded streets fall outside the regime the method was designed for. Semantic
discovery is performed per frame and does not assign persistent identities to instances across
frames, and the method does not produce confidence scores, which limits its use in score-based
evaluation protocols. Finally, the robust configuration costs about three seconds per frame on a
single L4 GPU; while the cascade adapts this cost to scene ambiguity, it is not real time.

\paragraph{Future work.}
Natural extensions are to link per-frame discoveries into temporally consistent instances by
reusing the same memory-based propagation employed for instance tracking, to derive
calibrated confidence scores from the embedding and VLM stages, and to distil the judgements of the
VLM into the embedding filter so that the uncertainty interval, and with it the latency, shrinks
over time.

\end{multicols}
\vspace{4mm}\noindent{\color{g200}\rule{\linewidth}{0.4pt}}
\begin{multicols}{2}

\end{multicols}


\begin{thebibliography}{99}

\bibitem{videolisa}
Zechen Bai, Tong He, Haiyang Mei, Pichao Wang, Ziteng Gao, Joya Chen, Lei Liu, Zheng Zhang, and Mike Zheng Shou.
\newblock One Token to Seg Them All: Language Instructed Reasoning Segmentation in Videos.
\newblock In NeurIPS, 2024.

\bibitem{qwen25vl}
Shuai Bai, Keqin Chen, Xuejing Liu, Jialin Wang, Wenbin Ge, Sibo Song, Kai Dang, Peng Wang, Shijie Wang, Jun Tang, Humen Zhong, Yuanzhi Zhu, Mingkun Yang, Zhaohai Li, Jianqiang Wan, Pengfei Wang, Wei Ding, Zheren Fu, Yiheng Xu, Jiabo Ye, Xi Zhang, Tianbao Xie, Zesen Cheng, Hang Zhang, Zhibo Yang, Haiyang Xu, and Junyang Lin.
\newblock Qwen2.5-VL Technical Report.
\newblock arXiv preprint arXiv:2502.13923, 2025.

\bibitem{qwen3vl}
Shuai Bai, Yuxuan Cai, Ruizhe Chen, Keqin Chen, Xionghui Chen, Zesen Cheng, Lianghao Deng, Wei Ding, Chang Gao, Chunjiang Ge, Wenbin Ge, Zhifang Guo, Qidong Huang, Jie Huang, Fei Huang, Binyuan Hui, Shutong Jiang, Zhaohai Li, Mingsheng Li, Mei Li, Kaixin Li, Zicheng Lin, Junyang Lin, Xuejing Liu, Jiawei Liu, Chenglong Liu, Yang Liu, Dayiheng Liu, Shixuan Liu, Dunjie Lu, Ruilin Luo, Chenxu Lv, Rui Men, Lingchen Meng, Xuancheng Ren, Xingzhang Ren, Sibo Song, Yuchong Sun, Jun Tang, Jianhong Tu, Jianqiang Wan, Peng Wang, Pengfei Wang, Qiuyue Wang, Yuxuan Wang, Tianbao Xie, Yiheng Xu, Haiyang Xu, Jin Xu, Zhibo Yang, Mingkun Yang, Jianxin Yang, An Yang, Bowen Yu, Fei Zhang, Hang Zhang, Xi Zhang, Bo Zheng, Humen Zhong, Jingren Zhou, Fan Zhou, Jing Zhou, Yuanzhi Zhu, and Ke Zhu.
\newblock Qwen3-VL Technical Report.
\newblock arXiv preprint arXiv:2511.21631, 2025.

\bibitem{paligemma}
Lucas Beyer, Andreas Steiner, Andr\'e Susano Pinto, Alexander Kolesnikov, Xiao Wang, Daniel Salz, Maxim Neumann, Ibrahim Alabdulmohsin, Michael Tschannen, Emanuele Bugliarello, Thomas Unterthiner, Daniel Keysers, Skanda Koppula, Fangyu Liu, Adam Grycner, Alexey Gritsenko, Neil Houlsby, Manoj Kumar, Keran Rong, Julian Eisenschlos, Rishabh Kabra, Matthias Bauer, Matko Bo\v{s}njak, Xi Chen, Matthias Minderer, Paul Voigtlaender, Ioana Bica, Ivana Balazevic, Joan Puigcerver, Pinelopi Papalampidi, Olivier Henaff, Xi Xiong, Radu Soricut, Jeremiah Harmsen, and Xiaohua Zhai.
\newblock PaliGemma: A versatile 3B VLM for transfer.
\newblock arXiv preprint arXiv:2407.07726, 2024.

\bibitem{detr}
Nicolas Carion, Francisco Massa, Gabriel Synnaeve, Nicolas Usunier, Alexander Kirillov, and Sergey Zagoruyko.
\newblock End-to-End Object Detection with Transformers.
\newblock In ECCV, 2020.

\bibitem{sam3}
Nicolas Carion, Laura Gustafson, Yuan-Ting Hu, Shoubhik Debnath, Ronghang Hu, Didac Suris, Chaitanya Ryali, Kalyan Vasudev Alwala, Haitham Khedr, Andrew Huang, Jie Lei, Tengyu Ma, Baishan Guo, Arpit Kalla, Markus Marks, Joseph Greer, Meng Wang, Peize Sun, Roman R\"adle, Triantafyllos Afouras, Effrosyni Mavroudi, Katherine Xu, Tsung-Han Wu, Yu Zhou, Liliane Momeni, Rishi Hazra, Shuangrui Ding, Sagar Vaze, Francois Porcher, Feng Li, Siyuan Li, Aishwarya Kamath, Ho Kei Cheng, Piotr Doll\'ar, Nikhila Ravi, Kate Saenko, Pengchuan Zhang, and Christoph Feichtenhofer.
\newblock SAM 3: Segment Anything with Concepts.
\newblock arXiv preprint arXiv:2511.16719, 2025.

\bibitem{saga}
Jiazhong Cen, Jiemin Fang, Chen Yang, Lingxi Xie, Xiaopeng Zhang, Wei Shen, and Qi Tian.
\newblock Segment Any 3D Gaussians.
\newblock In AAAI, 2025.

\bibitem{frugalgpt}
Lingjiao Chen, Matei Zaharia, and James Zou.
\newblock FrugalGPT: How to Use Large Language Models While Reducing Cost and Improving Performance.
\newblock arXiv preprint arXiv:2305.05176, 2023.

\bibitem{internvl}
Zhe Chen, Jiannan Wu, Wenhai Wang, Weijie Su, Guo Chen, Sen Xing, Muyan Zhong, Qinglong Zhang, Xizhou Zhu, Lewei Lu, Bin Li, Ping Luo, Tong Lu, Yu Qiao, and Jifeng Dai.
\newblock InternVL: Scaling up Vision Foundation Models and Aligning for Generic Visual-Linguistic Tasks.
\newblock In CVPR, 2024.

\bibitem{mask2former}
Bowen Cheng, Ishan Misra, Alexander G. Schwing, Alexander Kirillov, and Rohit Girdhar.
\newblock Masked-attention Mask Transformer for Universal Image Segmentation.
\newblock In CVPR, 2022.

\bibitem{xmem}
Ho Kei Cheng and Alexander G. Schwing.
\newblock XMem: Long-Term Video Object Segmentation with an Atkinson-Shiffrin Memory Model.
\newblock In ECCV, 2022.

\bibitem{cutie}
Ho Kei Cheng, Seoung Wug Oh, Brian Price, Joon-Young Lee, and Alexander Schwing.
\newblock Putting the Object Back into Video Object Segmentation.
\newblock In CVPR, 2024.

\bibitem{samtrack}
Yangming Cheng, Liulei Li, Yuanyou Xu, Xiaodi Li, Zongxin Yang, Wenguan Wang, and Yi Yang.
\newblock Segment and Track Anything.
\newblock arXiv preprint arXiv:2305.06558, 2023.

\bibitem{deva}
Ho Kei Cheng, Seoung Wug Oh, Brian Price, Alexander Schwing, and Joon-Young Lee.
\newblock Tracking Anything with Decoupled Video Segmentation.
\newblock In ICCV, 2023.

\bibitem{yoloworld}
Tianheng Cheng, Lin Song, Yixiao Ge, Wenyu Liu, Xinggang Wang, and Ying Shan.
\newblock YOLO-World: Real-Time Open-Vocabulary Object Detection.
\newblock In CVPR, 2024.

\bibitem{navit}
Mostafa Dehghani, Basil Mustafa, Josip Djolonga, Jonathan Heek, Matthias Minderer, Mathilde Caron, Andreas Steiner, Joan Puigcerver, Robert Geirhos, Ibrahim Alabdulmohsin, Avital Oliver, Piotr Padlewski, Alexey Gritsenko, Mario Lu\v{c}i\'c, and Neil Houlsby.
\newblock Patch n' Pack: NaViT, a Vision Transformer for any Aspect Ratio and Resolution.
\newblock In NeurIPS, 2023.

\bibitem{mose}
Henghui Ding, Chang Liu, Shuting He, Xudong Jiang, Philip H. S. Torr, and Song Bai.
\newblock MOSE: A New Dataset for Video Object Segmentation in Complex Scenes.
\newblock In ICCV, 2023.

\bibitem{mevis}
Henghui Ding, Chang Liu, Shuting He, Xudong Jiang, and Chen Change Loy.
\newblock MeViS: A Large-scale Benchmark for Video Segmentation with Motion Expressions.
\newblock In ICCV, 2023.

\bibitem{sam2long}
Shuangrui Ding, Rui Qian, Xiaoyi Dong, Pan Zhang, Yuhang Zang, Yuhang Cao, Yuwei Guo, Dahua Lin, and Jiaqi Wang.
\newblock SAM2Long: Enhancing SAM 2 for Long Video Segmentation with a Training-Free Memory Tree.
\newblock In ICCV, 2025.

\bibitem{selective}
Yonatan Geifman and Ran El-Yaniv.
\newblock Selective Classification for Deep Neural Networks.
\newblock In NeurIPS, 2017.

\bibitem{openseg}
Golnaz Ghiasi, Xiuye Gu, Yin Cui, and Tsung-Yi Lin.
\newblock Scaling Open-Vocabulary Image Segmentation with Image-Level Labels.
\newblock In ECCV, 2022.

\bibitem{vild}
Xiuye Gu, Tsung-Yi Lin, Weicheng Kuo, and Yin Cui.
\newblock Open-vocabulary Object Detection via Vision and Language Knowledge Distillation.
\newblock In ICLR, 2022.

\bibitem{calibration}
Chuan Guo, Geoff Pleiss, Yu Sun, and Kilian Q. Weinberger.
\newblock On Calibration of Modern Neural Networks.
\newblock In ICML, 2017.

\bibitem{lvis}
Agrim Gupta, Piotr Doll\'ar, and Ross Girshick.
\newblock LVIS: A Dataset for Large Vocabulary Instance Segmentation.
\newblock In CVPR, 2019.

\bibitem{maskrcnn}
Kaiming He, Georgia Gkioxari, Piotr Doll\'ar, and Ross Girshick.
\newblock Mask R-CNN.
\newblock In ICCV, 2017.

\bibitem{oneformer}
Jitesh Jain, Jiachen Li, MangTik Chiu, Ali Hassani, Nikita Orlov, and Humphrey Shi.
\newblock OneFormer: One Transformer to Rule Universal Image Segmentation.
\newblock In CVPR, 2023.

\bibitem{align}
Chao Jia, Yinfei Yang, Ye Xia, Yi-Ting Chen, Zarana Parekh, Hieu Pham, Quoc V. Le, Yunhsuan Sung, Zhen Li, and Tom Duerig.
\newblock Scaling Up Visual and Vision-Language Representation Learning With Noisy Text Supervision.
\newblock In ICML, 2021.

\bibitem{hqsam}
Lei Ke, Mingqiao Ye, Martin Danelljan, Yifan Liu, Yu-Wing Tai, Chi-Keung Tang, and Fisher Yu.
\newblock Segment Anything in High Quality.
\newblock In NeurIPS, 2023.

\bibitem{gs}
Bernhard Kerbl, Georgios Kopanas, Thomas Leimk\"uhler, and George Drettakis.
\newblock 3D Gaussian Splatting for Real-Time Radiance Field Rendering.
\newblock ACM Transactions on Graphics, 2023.

\bibitem{sam}
Alexander Kirillov, Eric Mintun, Nikhila Ravi, Hanzi Mao, Chloe Rolland, Laura Gustafson, Tete Xiao, Spencer Whitehead, Alexander C. Berg, Wan-Yen Lo, Piotr Doll\'ar, and Ross Girshick.
\newblock Segment Anything.
\newblock In ICCV, 2023.

\bibitem{wildgaussians}
Jonas Kulhanek, Songyou Peng, Zuzana Kukelova, Marc Pollefeys, and Torsten Sattler.
\newblock WildGaussians: 3D Gaussian Splatting in the Wild.
\newblock In NeurIPS, 2024.

\bibitem{lisa}
Xin Lai, Zhuotao Tian, Yukang Chen, Yanwei Li, Yuhui Yuan, Shu Liu, and Jiaya Jia.
\newblock LISA: Reasoning Segmentation via Large Language Model.
\newblock In CVPR, 2024.

\bibitem{glip}
Liunian Harold Li, Pengchuan Zhang, Haotian Zhang, Jianwei Yang, Chunyuan Li, Yiwu Zhong, Lijuan Wang, Lu Yuan, Lei Zhang, Jenq-Neng Hwang, Kai-Wei Chang, and Jianfeng Gao.
\newblock Grounded Language-Image Pre-training.
\newblock In CVPR, 2022.

\bibitem{lseg}
Boyi Li, Kilian Q. Weinberger, Serge Belongie, Vladlen Koltun, and Ren\'e Ranftl.
\newblock Language-driven Semantic Segmentation.
\newblock In ICLR, 2022.

\bibitem{teta}
Siyuan Li, Martin Danelljan, Henghui Ding, Thomas E. Huang, and Fisher Yu.
\newblock Tracking Every Thing in the Wild.
\newblock In ECCV, 2022.

\bibitem{blip2}
Junnan Li, Dongxu Li, Silvio Savarese, and Steven Hoi.
\newblock BLIP-2: Bootstrapping Language-Image Pre-training with Frozen Image Encoders and Large Language Models.
\newblock In ICML, 2023.

\bibitem{pope}
Yifan Li, Yifan Du, Kun Zhou, Jinpeng Wang, Wayne Xin Zhao, and Ji-Rong Wen.
\newblock Evaluating Object Hallucination in Large Vision-Language Models.
\newblock In EMNLP, 2023.

\bibitem{ovtrack}
Siyuan Li, Tobias Fischer, Lei Ke, Henghui Ding, Martin Danelljan, and Fisher Yu.
\newblock OVTrack: Open-Vocabulary Multiple Object Tracking.
\newblock In CVPR, 2023.

\bibitem{semanticsam}
Feng Li, Hao Zhang, Peize Sun, Xueyan Zou, Shilong Liu, Jianwei Yang, Chunyuan Li, Lei Zhang, and Jianfeng Gao.
\newblock Semantic-SAM: Segment and Recognize Anything at Any Granularity.
\newblock arXiv preprint arXiv:2307.04767, 2023.

\bibitem{coco}
Tsung-Yi Lin, Michael Maire, Serge Belongie, Lubomir Bourdev, Ross Girshick, James Hays, Pietro Perona, Deva Ramanan, C. Lawrence Zitnick, and Piotr Doll\'ar.
\newblock Microsoft COCO: Common Objects in Context.
\newblock In ECCV, 2014.

\bibitem{groundingdino}
Shilong Liu, Zhaoyang Zeng, Tianhe Ren, Feng Li, Hao Zhang, Jie Yang, Qing Jiang, Chunyuan Li, Jianwei Yang, Hang Su, Jun Zhu, and Lei Zhang.
\newblock Grounding DINO: Marrying DINO with Grounded Pre-Training for Open-Set Object Detection.
\newblock In ECCV, 2024.

\bibitem{llava}
Haotian Liu, Chunyuan Li, Qingyang Wu, and Yong Jae Lee.
\newblock Visual Instruction Tuning.
\newblock In NeurIPS, 2023.

\bibitem{hota}
Jonathon Luiten, Aljosa Osep, Patrick Dendorfer, Philip Torr, Andreas Geiger, Laura Leal-Taixe, and Bastian Leibe.
\newblock HOTA: A Higher Order Metric for Evaluating Multi-Object Tracking.
\newblock IJCV, 2021.

\bibitem{nerfw}
Ricardo Martin-Brualla, Noha Radwan, Mehdi S. M. Sajjadi, Jonathan T. Barron, Alexey Dosovitskiy, and Daniel Duckworth.
\newblock NeRF in the Wild: Neural Radiance Fields for Unconstrained Photo Collections.
\newblock In CVPR, 2021.

\bibitem{nerf}
Ben Mildenhall, Pratul P. Srinivasan, Matthew Tancik, Jonathan T. Barron, Ravi Ramamoorthi, and Ren Ng.
\newblock NeRF: Representing Scenes as Neural Radiance Fields for View Synthesis.
\newblock In ECCV, 2020.

\bibitem{owlvit}
Matthias Minderer, Alexey Gritsenko, Austin Stone, Maxim Neumann, Dirk Weissenborn, Alexey Dosovitskiy, Aravindh Mahendran, Anurag Arnab, Mostafa Dehghani, Zhuoran Shen, Xiao Wang, Xiaohua Zhai, Thomas Kipf, and Neil Houlsby.
\newblock Simple Open-Vocabulary Object Detection with Vision Transformers.
\newblock In ECCV, 2022.

\bibitem{owlv2}
Matthias Minderer, Alexey Gritsenko, and Neil Houlsby.
\newblock Scaling Open-Vocabulary Object Detection.
\newblock In NeurIPS, 2023.

\bibitem{stm}
Seoung Wug Oh, Joon-Young Lee, Ning Xu, and Seon Joo Kim.
\newblock Video Object Segmentation using Space-Time Memory Networks.
\newblock In ICCV, 2019.

\bibitem{davis}
Jordi Pont-Tuset, Federico Perazzi, Sergi Caelles, Pablo Arbel\'aez, Alex Sorkine-Hornung, and Luc Van Gool.
\newblock The 2017 DAVIS Challenge on Video Object Segmentation.
\newblock arXiv preprint arXiv:1704.00675, 2017.

\bibitem{ovis}
Jiyang Qi, Yan Gao, Yao Hu, Xinggang Wang, Xiaoyu Liu, Xiang Bai, Serge Belongie, Alan Yuille, Philip H. S. Torr, and Song Bai.
\newblock Occluded Video Instance Segmentation: A Benchmark.
\newblock IJCV, 2022.

\bibitem{langsplat}
Minghan Qin, Wanhua Li, Jiawei Zhou, Haoqian Wang, and Hanspeter Pfister.
\newblock LangSplat: 3D Language Gaussian Splatting.
\newblock In CVPR, 2024.

\bibitem{clip}
Alec Radford, Jong Wook Kim, Chris Hallacy, Aditya Ramesh, Gabriel Goh, Sandhini Agarwal, Girish Sastry, Amanda Askell, Pamela Mishkin, Jack Clark, Gretchen Krueger, and Ilya Sutskever.
\newblock Learning Transferable Visual Models From Natural Language Supervision.
\newblock In ICML, 2021.

\bibitem{glamm}
Hanoona Rasheed, Muhammad Maaz, Sahal Shaji Mullappilly, Abdelrahman Shaker, Salman Khan, Hisham Cholakkal, Rao M. Anwer, Erix Xing, Ming-Hsuan Yang, and Fahad S. Khan.
\newblock GLaMM: Pixel Grounding Large Multimodal Model.
\newblock In CVPR, 2024.

\bibitem{sam2}
Nikhila Ravi, Valentin Gabeur, Yuan-Ting Hu, Ronghang Hu, Chaitanya Ryali, Tengyu Ma, Haitham Khedr, Roman R\"adle, Chloe Rolland, Laura Gustafson, Eric Mintun, Junting Pan, Kalyan Vasudev Alwala, Nicolas Carion, Chao-Yuan Wu, Ross Girshick, Piotr Doll\'ar, and Christoph Feichtenhofer.
\newblock SAM 2: Segment Anything in Images and Videos.
\newblock In ICLR, 2025.

\bibitem{fasterrcnn}
Shaoqing Ren, Kaiming He, Ross Girshick, and Jian Sun.
\newblock Faster R-CNN: Towards Real-Time Object Detection with Region Proposal Networks.
\newblock In NeurIPS, 2015.

\bibitem{groundedsam}
Tianhe Ren, Shilong Liu, Ailing Zeng, Jing Lin, Kunchang Li, He Cao, Jiayu Chen, Xinyu Huang, Yukang Chen, Feng Yan, Zhaoyang Zeng, Hao Zhang, Feng Li, Jie Yang, Hongyang Li, Qing Jiang, and Lei Zhang.
\newblock Grounded SAM: Assembling Open-World Models for Diverse Visual Tasks.
\newblock arXiv preprint arXiv:2401.14159, 2024.

\bibitem{groundingdino15}
Tianhe Ren, Qing Jiang, Shilong Liu, Zhaoyang Zeng, Wenlong Liu, Han Gao, Hongjie Huang, Zhengyu Ma, Xiaoke Jiang, Yihao Chen, Yuda Xiong, Hao Zhang, Feng Li, Peijun Tang, Kent Yu, and Lei Zhang.
\newblock Grounding DINO 1.5: Advance the ''Edge'' of Open-Set Object Detection.
\newblock arXiv preprint arXiv:2405.10300, 2024.

\bibitem{nerfonthego}
Weining Ren, Zihan Zhu, Boyang Sun, Jiaqi Chen, Marc Pollefeys, and Songyou Peng.
\newblock NeRF On-the-go: Exploiting Uncertainty for Distractor-free NeRFs in the Wild.
\newblock In CVPR, 2024.

\bibitem{hiera}
Chaitanya Ryali, Yuan-Ting Hu, Daniel Bolya, Chen Wei, Haoqi Fan, Po-Yao Huang, Vaibhav Aggarwal, Arkabandhu Chowdhury, Omid Poursaeed, Judy Hoffman, Jitendra Malik, Yanghao Li, and Christoph Feichtenhofer.
\newblock Hiera: A Hierarchical Vision Transformer without the Bells-and-Whistles.
\newblock In ICML, 2023.

\bibitem{robustnerf}
Sara Sabour, Suhani Vora, Daniel Duckworth, Ivan Krasin, David J. Fleet, and Andrea Tagliasacchi.
\newblock RobustNeRF: Ignoring Distractors with Robust Losses.
\newblock In CVPR, 2023.

\bibitem{spotless}
Sara Sabour, Lily Goli, George Kopanas, Mark Matthews, Dmitry Lagun, Leonidas Guibas, Alec Jacobson, David J. Fleet, and Andrea Tagliasacchi.
\newblock SpotlessSplats: Ignoring Distractors in 3D Gaussian Splatting.
\newblock arXiv preprint arXiv:2406.20055, 2024.

\bibitem{laion5b}
Christoph Schuhmann, Romain Beaumont, Richard Vencu, Cade Gordon, Ross Wightman, Mehdi Cherti, Theo Coombes, Aarush Katta, Clayton Mullis, Mitchell Wortsman, Patrick Schramowski, Srivatsa Kundurthy, Katherine Crowson, Ludwig Schmidt, Robert Kaczmarczyk, and Jenia Jitsev.
\newblock LAION-5B: An open large-scale dataset for training next generation image-text models.
\newblock In NeurIPS, 2022.

\bibitem{siglip2}
Michael Tschannen, Alexey Gritsenko, Xiao Wang, Muhammad Ferjad Naeem, Ibrahim Alabdulmohsin, Nikhil Parthasarathy, Talfan Evans, Lucas Beyer, Ye Xia, Basil Mustafa, Olivier H\'enaff, Jeremiah Harmsen, Andreas Steiner, and Xiaohua Zhai.
\newblock SigLIP 2: Multilingual Vision-Language Encoders with Improved Semantic Understanding, Localization, and Dense Features.
\newblock arXiv preprint arXiv:2502.14786, 2025.

\bibitem{dam4sam}
Jovana Videnovic, Alan Lukezic, and Matej Kristan.
\newblock A Distractor-Aware Memory for Visual Object Tracking with SAM2.
\newblock In CVPR, 2025.

\bibitem{qwen2vl}
Peng Wang, Shuai Bai, Sinan Tan, Shijie Wang, Zhihao Fan, Jinze Bai, Keqin Chen, Xuejing Liu, Jialin Wang, Wenbin Ge, Yang Fan, Kai Dang, Mengfei Du, Xuancheng Ren, Rui Men, Dayiheng Liu, Chang Zhou, Jingren Zhou, and Junyang Lin.
\newblock Qwen2-VL: Enhancing Vision-Language Model's Perception of the World at Any Resolution.
\newblock arXiv preprint arXiv:2409.12191, 2024.

\bibitem{cot}
Jason Wei, Xuezhi Wang, Dale Schuurmans, Maarten Bosma, Brian Ichter, Fei Xia, Ed Chi, Quoc Le, and Denny Zhou.
\newblock Chain-of-Thought Prompting Elicits Reasoning in Large Language Models.
\newblock In NeurIPS, 2022.

\bibitem{outlines}
Brandon T. Willard and R\'emi Louf.
\newblock Efficient Guided Generation for Large Language Models.
\newblock arXiv preprint arXiv:2307.09702, 2023.

\bibitem{referformer}
Jiannan Wu, Yi Jiang, Peize Sun, Zehuan Yuan, and Ping Luo.
\newblock Language as Queries for Referring Video Object Segmentation.
\newblock In CVPR, 2022.

\bibitem{florence2}
Bin Xiao, Haiping Wu, Weijian Xu, Xiyang Dai, Houdong Hu, Yumao Lu, Michael Zeng, Ce Liu, and Lu Yuan.
\newblock Florence-2: Advancing a Unified Representation for a Variety of Vision Tasks.
\newblock In CVPR, 2024.

\bibitem{efficientsam}
Yunyang Xiong, Bala Varadarajan, Lemeng Wu, Xiaoyu Xiang, Fanyi Xiao, Chenchen Zhu, Xiaoliang Dai, Dilin Wang, Fei Sun, Forrest Iandola, Raghuraman Krishnamoorthi, and Vikas Chandra.
\newblock EfficientSAM: Leveraged Masked Image Pretraining for Efficient Segment Anything.
\newblock In CVPR, 2024.

\bibitem{ytvos}
Ning Xu, Linjie Yang, Yuchen Fan, Dingcheng Yue, Yuchen Liang, Jianchao Yang, and Thomas Huang.
\newblock YouTube-VOS: A Large-Scale Video Object Segmentation Benchmark.
\newblock arXiv preprint arXiv:1809.03327, 2018.

\bibitem{odise}
Jiarui Xu, Sifei Liu, Arash Vahdat, Wonmin Byeon, Xiaolong Wang, and Shalini De Mello.
\newblock Open-Vocabulary Panoptic Segmentation with Text-to-Image Diffusion Models.
\newblock In CVPR, 2023.

\bibitem{san}
Mengde Xu, Zheng Zhang, Fangyun Wei, Han Hu, and Xiang Bai.
\newblock Side Adapter Network for Open-Vocabulary Semantic Segmentation.
\newblock In CVPR, 2023.

\bibitem{visa}
Cilin Yan, Haochen Wang, Shilin Yan, Xiaolong Jiang, Yao Hu, Guoliang Kang, Weidi Xie, and Efstratios Gavves.
\newblock VISA: Reasoning Video Object Segmentation via Large Language Models.
\newblock In ECCV, 2024.

\bibitem{lavt}
Zhao Yang, Jiaqi Wang, Yansong Tang, Kai Chen, Hengshuang Zhao, and Philip H. S. Torr.
\newblock LAVT: Language-Aware Vision Transformer for Referring Image Segmentation.
\newblock In CVPR, 2022.

\bibitem{deaot}
Zongxin Yang and Yi Yang.
\newblock Decoupling Features in Hierarchical Propagation for Video Object Segmentation.
\newblock In NeurIPS, 2022.

\bibitem{samurai}
Cheng-Yen Yang, Hsiang-Wei Huang, Wenhao Chai, Zhongyu Jiang, and Jenq-Neng Hwang.
\newblock SAMURAI: Adapting Segment Anything Model for Zero-Shot Visual Tracking with Motion-Aware Memory.
\newblock arXiv preprint arXiv:2411.11922, 2024.

\bibitem{gaussiangrouping}
Mingqiao Ye, Martin Danelljan, Fisher Yu, and Lei Ke.
\newblock Gaussian Grouping: Segment and Edit Anything in 3D Scenes.
\newblock In ECCV, 2024.

\bibitem{florence}
Lu Yuan, Dongdong Chen, Yi-Ling Chen, Noel Codella, Xiyang Dai, Jianfeng Gao, Houdong Hu, Xuedong Huang, Boxin Li, Chunyuan Li, Ce Liu, Mengchen Liu, Zicheng Liu, Yumao Lu, Yu Shi, Lijuan Wang, Jianfeng Wang, Bin Xiao, Zhen Xiao, Jianwei Yang, Michael Zeng, Luowei Zhou, and Pengchuan Zhang.
\newblock Florence: A New Foundation Model for Computer Vision.
\newblock arXiv preprint arXiv:2111.11432, 2021.

\bibitem{sa2va}
Haobo Yuan, Xiangtai Li, Tao Zhang, Yueyi Sun, Zilong Huang, Shilin Xu, Shunping Ji, Yunhai Tong, Lu Qi, Jiashi Feng, and Ming-Hsuan Yang.
\newblock Sa2VA: Marrying SAM2 with MLLM for Dense Grounded Understanding of Images and Videos.
\newblock IEEE TPAMI, 2025.

\bibitem{siglip}
Xiaohua Zhai, Basil Mustafa, Alexander Kolesnikov, and Lucas Beyer.
\newblock Sigmoid Loss for Language Image Pre-Training.
\newblock In ICCV, 2023.

\bibitem{bytetrack}
Yifu Zhang, Peize Sun, Yi Jiang, Dongdong Yu, Fucheng Weng, Zehuan Yuan, Ping Luo, Wenyu Liu, and Xinggang Wang.
\newblock ByteTrack: Multi-Object Tracking by Associating Every Detection Box.
\newblock In ECCV, 2022.

\bibitem{dinodet}
Hao Zhang, Feng Li, Shilong Liu, Lei Zhang, Hang Su, Jun Zhu, Lionel M. Ni, and Heung-Yeung Shum.
\newblock DINO: DETR with Improved DeNoising Anchor Boxes for End-to-End Object Detection.
\newblock In ICLR, 2023.

\bibitem{openseed}
Hao Zhang, Feng Li, Xueyan Zou, Shilong Liu, Chunyuan Li, Jianfeng Gao, Jianwei Yang, and Lei Zhang.
\newblock A Simple Framework for Open-Vocabulary Segmentation and Detection.
\newblock In ICCV, 2023.

\bibitem{sec}
Zhixiong Zhang, Shuangrui Ding, Xiaoyi Dong, Songxin He, Jianfan Lin, Junsong Tang, Yuhang Zang, Yuhang Cao, Dahua Lin, and Jiaqi Wang.
\newblock Advancing Complex Video Object Segmentation via Progressive Concept Construction.
\newblock arXiv preprint arXiv:2507.15852, 2025.

\bibitem{llmjudge}
Lianmin Zheng, Wei-Lin Chiang, Ying Sheng, Siyuan Zhuang, Zhanghao Wu, Yonghao Zhuang, Zi Lin, Zhuohan Li, Dacheng Li, Eric P. Xing, Hao Zhang, Joseph E. Gonzalez, and Ion Stoica.
\newblock Judging LLM-as-a-Judge with MT-Bench and Chatbot Arena.
\newblock In NeurIPS, 2023.

\bibitem{coop}
Kaiyang Zhou, Jingkang Yang, Chen Change Loy, and Ziwei Liu.
\newblock Learning to Prompt for Vision-Language Models.
\newblock IJCV, 2022.

\bibitem{detic}
Xingyi Zhou, Rohit Girdhar, Armand Joulin, Philipp Kr\"ahenb\"uhl, and Ishan Misra.
\newblock Detecting Twenty-thousand Classes using Image-level Supervision.
\newblock In ECCV, 2022.

\bibitem{xdecoder}
Xueyan Zou, Zi-Yi Dou, Jianwei Yang, Zhe Gan, Linjie Li, Chunyuan Li, Xiyang Dai, Harkirat Behl, Jianfeng Wang, Lu Yuan, Nanyun Peng, Lijuan Wang, Yong Jae Lee, and Jianfeng Gao.
\newblock Generalized Decoding for Pixel, Image, and Language.
\newblock In CVPR, 2023.

\bibitem{seem}
Xueyan Zou, Jianwei Yang, Hao Zhang, Feng Li, Linjie Li, Jianfeng Wang, Lijuan Wang, Jianfeng Gao, and Yong Jae Lee.
\newblock Segment Everything Everywhere All at Once.
\newblock In NeurIPS, 2023.

\balance
\end{thebibliography}
\end{document}